\documentclass[conference]{IEEEtran}

\usepackage{times}

\usepackage[numbers]{natbib}
\usepackage{multicol}
\usepackage[bookmarks=true, allcolors=blue]{hyperref}

\let\labelindent\relax
\usepackage{enumitem}
\usepackage{subfiles}
\usepackage{graphicx}
\usepackage{caption}
\usepackage{bm}
\usepackage{xcolor}
\usepackage{xspace}
\usepackage{amsmath}
\usepackage{amssymb}
\usepackage{multirow}
\usepackage{algorithm}
\usepackage{algorithmicx}
\usepackage[free-standing-units=true]{siunitx}
\usepackage[noend]{algpseudocode}
\def\NoNumber#1{{\def\alglinenumber##1{}\State #1}\addtocounter{ALG@line}{-1}}
\ifCLASSOPTIONcompsoc
  \usepackage[caption=false,font=normalsize,labelfont=sf,textfont=sf]{subfig}
\else
  \usepackage[caption=false,font=footnotesize]{subfig}
\fi
\usepackage{balance}

\newcommand*{\Rn}{\ensuremath{\mathbb{R}^n}}
\newcommand*{\xstar}{\ensuremath{\bm x^\ast}}
\newcommand*{\Aone}{\ensuremath{\mathcal{A}_1}}

\newcommand*{\Atwo}{\ensuremath{\mathcal{A}_2}}
\newcommand*{\dshapes}{\ensuremath{\text{dist}(\Aone, \Atwo)}}
\newcommand*{\Ddiff}{\ensuremath{\mathcal{D}}}
\newcommand*{\Cspace}{\ensuremath{\mathcal{C}}}
\newcommand{\comment}[1]{}

\DeclareMathOperator*{\argmin}{arg\,min}
\algnewcommand\algorithmicswitch{\textbf{switch}}
\algnewcommand\algorithmiccase{\textbf{case}}
\algdef{SE}[SWITCH]{Switch}{EndSwitch}[1]{\algorithmicswitch\ #1\ \algorithmicdo}{\algorithmicend\ \algorithmicswitch}%
\algdef{SE}[CASE]{Case}{EndCase}[1]{\algorithmiccase\ #1}{\algorithmicend\ \algorithmiccase}%
\algtext*{EndSwitch}%
\algtext*{EndCase}%
\newcommand{\norm}[1]{\left\lVert#1\right\rVert}
\def\numit{\texttt{N}^{\texttt{k}}}

\def\exectime{\texttt{T}^{\mu}}

\newcommand*{\numitrel}{\ensuremath{\numit_{\text{GJK}} / \numit_{\text{Nesterov}}}}
\newcommand*{\exectimerel}{\ensuremath{\exectime_{\text{GJK}}/\exectime_{\text{Nesterov}}}}
\newcommand*{\exectimeD}{\ensuremath{\exectime_{\text{D}}}}
\newcommand*{\exectimeC}{\ensuremath{\exectime_{\text{C}}}}

\makeatletter
\DeclareRobustCommand\onedot{\futurelet\@let@token\@onedot}
\def\@onedot{\ifx\@let@token.\else.\null\fi\xspace}

\def\ie{\emph{i.e}\onedot}

\makeatother

\begin{document}

\title{Collision Detection Accelerated: \\ An Optimization Perspective}

\author{\authorblockN{Louis Montaut\authorrefmark{1}\authorrefmark{2},
Quentin Le Lidec\authorrefmark{2},
Vladimir Petrik\authorrefmark{1}, 
Josef Sivic\authorrefmark{1} and
Justin Carpentier\authorrefmark{2}}
\authorblockA{\authorrefmark{1}Czech Institute of Informatics, Robotics and Cybernetics,\\
        Czech Technical University, Prague, Czech Republic\\
        {\tt\small firstname.lastname@cvut.cz}
    }
\authorblockA{\authorrefmark{2}Inria and Département d'Informatique de l'\'Ecole Normale Supérieure,\\
        PSL Research University, Paris, France\\
        {\tt\small firstname.lastname@inria.fr}
    }
}

\maketitle

\begin{abstract}
    Collision detection between two convex shapes is an essential feature of any physics engine or robot motion planner.
It has often been  tackled as a computational geometry problem, with the Gilbert, Johnson and Keerthi~(GJK) algorithm being the most common approach today. 
In this work we leverage the fact that collision detection is fundamentally a convex optimization problem.
In particular, we establish that the GJK algorithm is a specific sub-case of the well-established Frank-Wolfe~(FW) algorithm in convex optimization.
We introduce a new collision detection algorithm by adapting recent works linking Nesterov acceleration and Frank-Wolfe methods.
We benchmark the proposed accelerated collision detection method on two  datasets composed of strictly convex and non-strictly convex shapes.
Our results show that our approach significantly reduces the number of iterations to solve collision detection problems compared to the state-of-the-art GJK algorithm, leading to up to two times faster computation times.

\end{abstract}

\IEEEpeerreviewmaketitle

% SECTION I
\section{Introduction}
\label{sec:introduction}

% INTRODUCTION
Physics engines designed to simulate rigid bodies are an essential tool used in a wide variety of applications, notably in robotics, video games, and computer graphics~\cite{coumans2021,physxweb,todorov2012mujoco}.
Collision detection, a crucial feature of any physics engine or robot motion planer~\cite{mirabel2016hpp,toussaint2018differentiable,schulman2014motion}, consists of finding which objects are colliding or not, \ie are sharing at least one common point or if there exists a separating hyper-plane between both.
As simulation often needs to deal with multiple objects and run in real-time (\ie, in video games) or at very high-frequencies (\ie, in robotics), it is crucial for collision detection to be carried out as fast as possible.
In order to reduce computational times, collision detection is usually decomposed into two phases thoroughly covered in~\cite{ericsonRealTimeCollisionDetection}.
The first phase is the \textit{broad phase} which consists in identifying which pair of simulated objects are potentially colliding.
Objects in the pair selected during the {broad phase} are close to each other, hence the collision is uncertain and needs to be considered carefully. %asserted or discarded carefully.
The second phase is the \textit{narrow phase} in which each pair identified in the broad phase is tested to check whether a collision is indeed occurring.
Collision detection during the narrow phase is the focus of this paper.
%%%%%%%%%% FIGURE %%%%%%%%%%
%%%%%%%%%%%%%%%%%%%%%%%%%%%%
\begin{figure}[!t]
    \centering
    \subfloat[\textbf{Left:} two distant shapes.
    \textbf{Right:} $\bm 0_{\Cspace}$ lies outside of the Minkowski difference of the shapes.
    \label{fig:shapes_mink_dist_a}]{
        \includegraphics[width=0.45\linewidth]{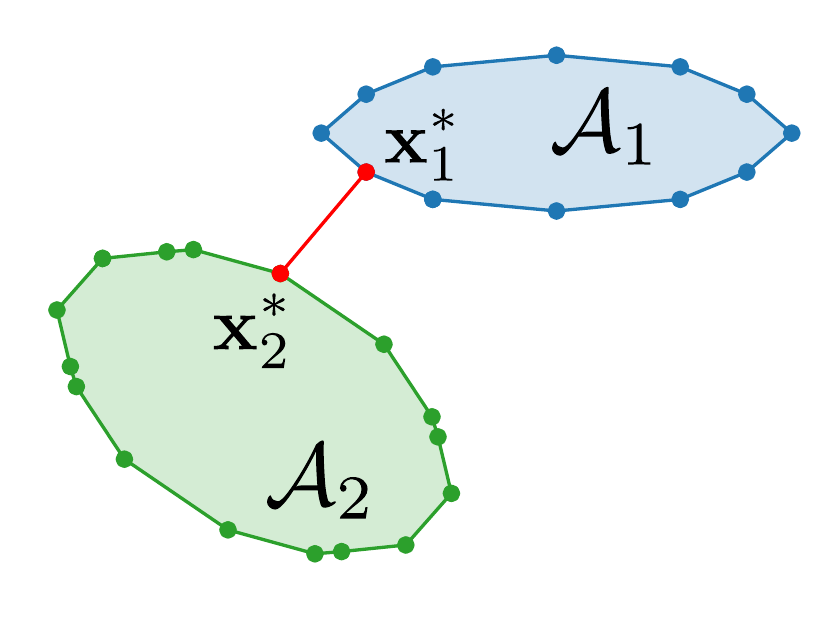}
        \hfill
        \includegraphics[width=0.45\linewidth]{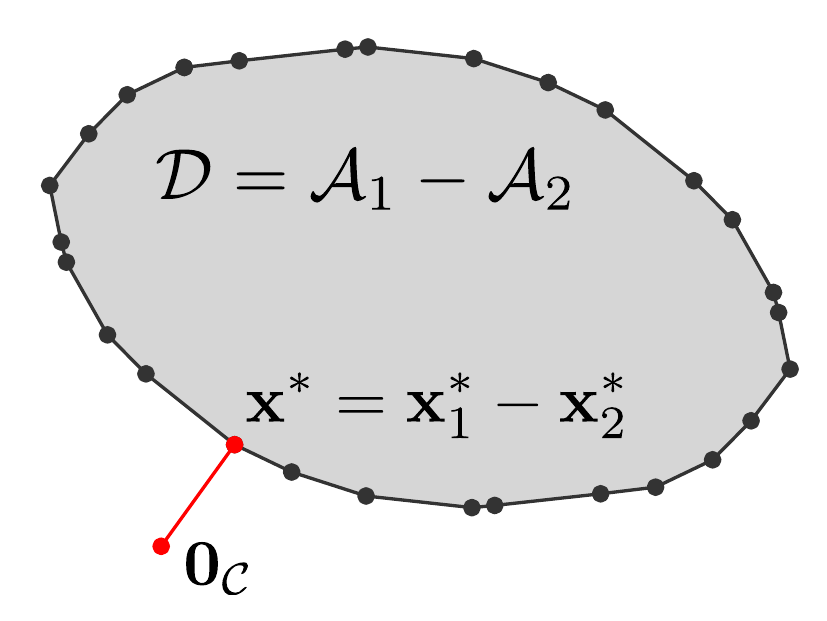}
    } \\
    \subfloat[\textbf{Left:} two overlapping shapes.
    \textbf{Right:} $\bm 0_{\Cspace}$ lies inside of the Minkowski difference of the shapes.
    \label{fig:shapes_mink_dist_b}]{
        \includegraphics[width=0.45\linewidth]{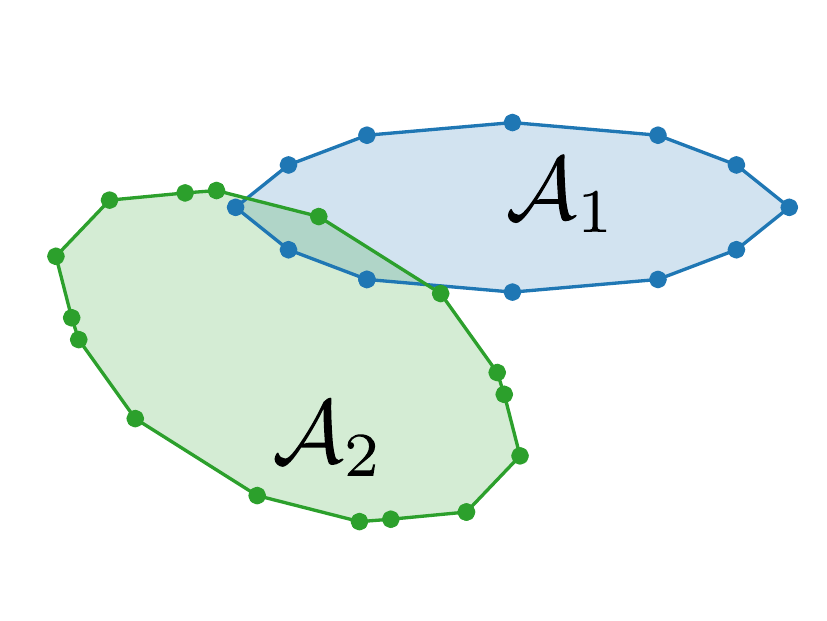}
        \hfill
        \includegraphics[width=0.45\linewidth]{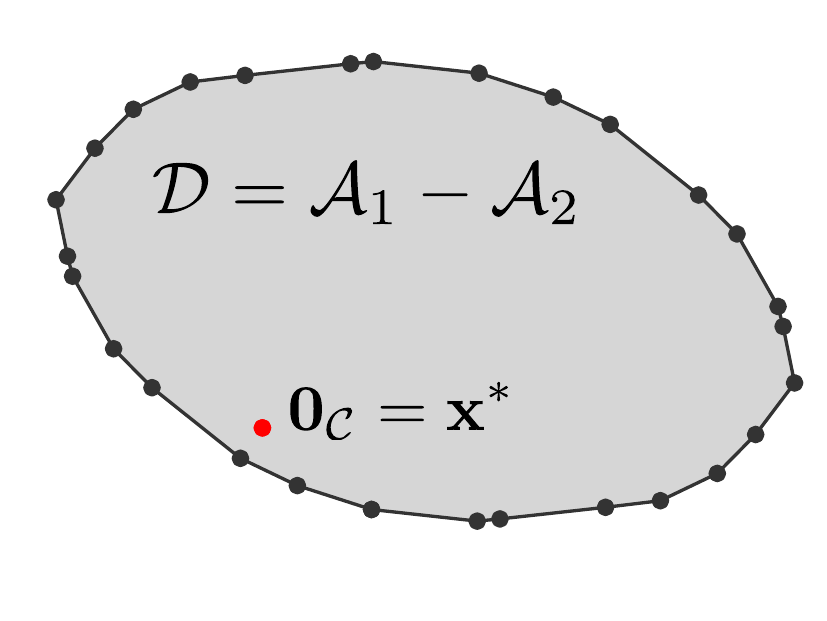}
    } 
    \caption{\small\textbf{Distant vs. overlapping pairs of shapes and their respective Minkowski difference.}
    Left column: two convex shapes in 2D.
    Right column: the Minkowski difference~$\Ddiff$ of~$\Aone$ and~$\Atwo$.
    Since~$\Aone$ and~$\Atwo$ are convex,~$\Ddiff$ is also convex.
    In (a), the shapes are not in collision hence the origin of the configuration space~$\Cspace$,~$\bm 0_{\Cspace}$ (in red) lies outside the Minkowski difference,~$\bm 0_{\Cspace} \not\in \Ddiff$.
    The vector~$\xstar = \bm x_1^* - \bm x_2^*$ separates~$\Aone$ from~$\Atwo$.
    It is also equal to the projection of~$\bm 0_{\Cspace}$ onto the Minkowski difference~$\Ddiff$,~$\xstar = \text{proj}_{\Ddiff}(\bm 0_{\Cspace})$.
    In (b), the shapes are overlapping, thus~$\bm 0_{\Cspace} \in \Ddiff$.
    In this case, we have~$\xstar = \text{proj}_{\Ddiff}(\bm 0_{\Cspace}) = \bm 0_{\Cspace}$.
    }
    \label{fig:shapes_mink_dist}
    \vspace{-0.5cm}
\end{figure}
%%%%%%%%%%%%%%%%%%%%%%%%%%%%
%%%%%%%%%%%%%%%%%%%%%%%%%%%%

% PROBLEM FORMULATION
\vspace{0.1cm}
\noindent
\textbf{Problem formulation}.
We consider two convex shapes~$\Aone$ and~$\Atwo$ in~$\mathbb{R}^n$~(with $n = 2$ or $3$ in common applications). 
If the shapes are not convex, we use their respective convex-hulls or decompose the shapes into a collection of convex sub-shapes~\cite{mamouSimpleEfficientApproach2009}.
The~\textit{separation distance} between~$\Aone$ and~$\Atwo$, denoted by~$\dshapes \in \mathbb{R}_{+}$, can be formulated as a minimization problem of the form:
\begin{equation}
    \begin{aligned}
         &d_{1,2} = \min_{ \bm x_1 \in \Aone, \bm x_2 \in \Atwo} \norm{\bm x_1 - \bm x_2 }^2\\
         &\text{and }
         \dshapes = \sqrt{d_{1,2}} 
        \,,
    \end{aligned}
    \label{eq:distance_computation_problem_shapes}
\end{equation}
%%% --> END EQUATION
where~$\bm x_1 \in \Aone$ and~$\bm x_2 \in \Atwo$ are both vectors in~$\mathbb{R}^n$,~$d_{1,2}$ is the optimal value of~\eqref{eq:distance_computation_problem_shapes} and~$\norm{\cdot}$ is the Euclidian norm of~$\Rn$.
If~$\Aone$ and~$\Atwo$ are intersecting (\ie in collision), we have~$\dshapes = 0$.
If the two shapes do not intersect, we have ~$\dshapes > 0$.
These two cases are illustrated in Fig.~\ref{fig:shapes_mink_dist}.

Problem~\eqref{eq:distance_computation_problem_shapes} allows us to encapsulate and tackle both the~\textit{distance computation} problem and the computationally cheaper \textit{Boolean collision check} into one single convex optimization problem.
In the distance computation problem, we aim at computing the separation distance between~$\Aone$ and~$\Atwo$, denoted $\dshapes$, \ie the distance between their closest points.
This distance is useful in some applications such as collision-free path planning~\cite{gilbertDistanceFunctionsTheir1985,stasseRealtimeSelfCollision2008}, especially for pairs of objects entering the narrow phase.
If a pair of objects has not been selected by the broad phase, a cheap estimate of~$\dshapes$ is usually enough~\cite{ericsonRealTimeCollisionDetection}.
In the Boolean collision check we only aim at determining if~$\Aone$ and~$\Atwo$ intersect, and computing~$\dshapes$ is unnecessary.
However, we will later see that the Boolean collision check is a sub-problem of the distance computation problem: solving~\eqref{eq:distance_computation_problem_shapes} can be early-stopped once a separating plane between~$\Aone$ and~$\Atwo$ has been found.
In the rest of this paper, we will use the generic term ``collision detection'' when we mean to encapsulate both distance computation and Boolean collision checking.
We will specify when the distinction is needed.

%%%%%%%%%%%%%%%%%%%%%%%%
%%%%%%%%%%%%%%%%%%%%%%%%
%%%%%%%%%%%%%%%%%%%%%%%%
% RELATED WORKS
\vspace{0.1cm}
\noindent
\textbf{Related work}.
The most well-known algorithm for collision detection between two convex shapes is the so-called Gilbert-Johnson-Keerthi algorithm~(GJK)~\cite{GJK88}.
It can handle both the distance computation and the Boolean collision check~\cite{bergenFastRobustGJK1999}.
Most alternatives to GJK found in the literature focus on computing collisions between convex polyhedra such as the Lin-Canny algorithm~\cite{linFastAlgorithmIncremental1991} or the V-Clip~\cite{mirtichVClipFastRobust1998} algorithm.
Although GJK is equivalent in performance to these algorithms~\cite{cameronComparisonTwoFast1997}, it is not restricted to convex polyhedra.
The strength of GJK is formulating the collision detection problem on the Minkowski difference.
The properties of the Minkowski difference are used to cleverly compute support vectors on the Minkowski difference (these notions will be introduced in detail in Sec.~\ref{sec:frank_wolfe}).
GJK is thus able to handle collision detection and distance computation for many different shapes such as convex polyhedra and basic primitives~(\ie, spheres, ellipsoids, cylinders, capsules etc.)~\cite{bergenFastRobustGJK1999,vanDenBergenCollisionDetectionBook,ericsonRealTimeCollisionDetection}.
The generality of GJK as well as its efficiency, good precision and ease of implementation makes it the state-of-the-art algorithm for collision detection between two convex shapes.
The computation time of GJK has been reduced thanks to improvements to its sub-operations~\cite{montanariImprovingGJKAlgorithm2017,bergenFastRobustGJK1999}.
However, to the best of our knowledge, no previous work has focused explicitly on lowering the number of iterations of GJK.
This work notably shows how to accelerate collision detection by directly lowering the number of iterations needed to solve a collision problem instance compared to the vanilla GJK algorithm.

Over the years, collision detection has often been seen as a computational geometry problem.
However we argue that this view has limited the improvement of collision detection and propose to reframe it as a convex optimization problem.
As briefly mentioned already in their 1988 paper~\cite{GJK88} and brought-up again by~\cite{nesmino2019}, the ideas developed by Gilbert, Johnson and Keerthi are rooted in convex optimization, notably in the work of~\citet{wolfeFindingNearestPoint1976a} and~\citet{gilbertIterativeProcedureComputing1966}.
Similarly to GJK, these works aim at solving a Minimum-Norm Point problem and are sub-cases of a more general algorithm: the Frank-Wolfe method, also known as the conditional gradient descent.

The Frank-Wolfe algorithm (FW) dates back to 1956 and is one of the first convex optimization algorithms. 
It has been heavily studied over the years by the optimization community.
This algorithm iterates over the computation of~\textit{support points} in order to approach the optimal solution.
The undesired zig-zagging behavior of FW, already identified by its authors, has been addressed by introducing corrections to the original FW method~\cite{garberPlayingNonlinearGames2013,gilbertIterativeProcedureComputing1966,guelatCommentsWolfeAway1986,jaggi13,kerdreuxProjectionFreeOptimizationUniformly2020,lacoste-julienGlobalLinearConvergence2015a,wolfeFindingNearestPoint1976a}.
In~\cite{jaggi13} and~\cite{lacoste-julienGlobalLinearConvergence2015a}, widely used corrections of the FW algorithm are analyzed as well as their convergence properties.
In this work, we notably show in Sec.~\ref{sec:frank_wolfe} that the GJK algorithm is an instance of the \textit{fully-corrective} Frank-Wolfe algorithm, covered in~\cite{lacoste-julienGlobalLinearConvergence2015a}, applied to solving a Minimum-Norm Point problem (MNP).
Finally, recent works have also tried to accelerate the FW algorithm by applying the so-called Nesterov acceleration~\cite{nesterov1983AMF}, a classic acceleration technique in unconstrained optimization.
Nesterov momentum has been successfully added by~\citet{liMomentumGuidedFrankWolfeAlgorithm2021a} in order to accelerate FW.
In~\cite{nesmino2019}, Qin and An take a different approach as they are interested in the general problem of projecting a point onto a Minkowski difference in \textit{any} dimension.
In order to accelerate the theoretical convergence of the 1966 Gilbert algorithm, the authors devise the NESMINO algorithm which exploits the classic Nesterov acceleration.
However, the minimization problem~\eqref{eq:distance_computation_problem_shapes} is modified by introducing a smoothing term.
By doing so, the authors rely on successive projections onto the original shapes instead of computing support points.
This makes the NESMINO algorithm similar to the projected-gradient descent method.
Unlike NESMINO, GJK and our work are sub-cases of the FW algorithm.
As pointed out by the authors, although the NESMINO algorithm
makes use of the Nesterov acceleration, it does not accelerate over the original 1966 Gilbert algorithm.
In Sec.~\ref{sec:experiments}, we experimentally show that the NESMINO algorithm is slower when compared to GJK and our method.

%%%%%%%%%%%%%%
% CONTRIBUTIONS
\vspace{0.1cm}
\noindent
\textbf{Contributions.}
Our work builds on the seminal works by~\citet{frankAlgorithmQuadraticProgramming1956} and \citet{GJK88} as well as on the work of~\citet{liMomentumGuidedFrankWolfeAlgorithm2021a} to globally accelerate distance computation and collision checking algorithms between convex shapes.
We make these three main contributions:
\begin{itemize}[label=$\hookrightarrow$]
    \item We recast the collision detection problem as a convex optimization problem which can be solved by the FW algorithm. Using the ideas developed by Gilbert, Johnson and Keerthi, we show that GJK is in fact a sub-case of the fully-corrective FW algorithm;
    \item We adapt recent works on Nesterov-accelerated FW in order to accelerate both the distance computation and the Boolean collision check problems;
    \item We empirically analyse the convergence of our proposed approach on two large shape benchmarks.
        Results show a faster convergence of our approach leading to a computational time up to two times faster than the state-of-the-art GJK algorithm, on both distance computation and Boolean collision checking.
\end{itemize}

% PLAN
\noindent \textbf{Paper outline.}
The paper is organized as follows.
In Sec.~\ref{sec:frank_wolfe}, we recast the distance computation problem as a Frank-Wolfe instance.
We introduce the duality-gap of the FW method, allowing us to bound the distance to the optimal solution of the distance computation problem.
We also present the fully-corrective version of FW and show the link between GJK and FW.
In Sec.~\ref{sec:accelerated_collision_detection}, we introduce recent work on Nesterov-accelerated FW and show how to adapt it for both distance computation and Boolean collision checking.
For distance computation, we adapt the convergence criterion of FW when using Nesterov acceleration in order to retain the bound on the distance to the optimal solution.
We also propose to adapt the Nesterov acceleration for non-strictly convex shapes.
In Sec.~\ref{sec:experiments}, we evaluate our approach against the state-of-the-art GJK algorithm on two benchmarks containing both strictly convex shapes and non-strictly convex shapes.

% --> END SECTION I

% SECTION II
\section{Collision Detection from a Frank-Wolfe Perspective}
\label{sec:frank_wolfe}

In this section, we highlight the natural connection between computing the distance between convex shapes and convex optimization, particularly within the frame of the Frank-Wolfe setting. 
We notably show that the GJK algorithm can be seen as a variant of the Frank-Wolfe algorithm that leverages properties of convex 3D shapes to drastically lower the computational complexity. \\

%%%%%%%%%%%%%%%%%%%%%%
%%%%%%%%%%%%%%%%%%%%%%
%% DISTANCE COMPUTATION VS. BOOLEAN COLLISION CHECKING
\noindent \textbf{Distance computation and Boolean collision checking.}
As recalled in Sec.~\ref{sec:introduction}, collision detection is a sub-case of distance computation: $\dshapes > 0$ means that the two shapes do not overlap while $\dshapes = 0$ means that the shapes are in collision.
In the particular case of $\dshapes > 0$, it is sufficient to find a strictly positive lower bound on~$d_{1,2}$ to solve the collision problem.
In the context of convex shapes, this is often simpler than computing the distance between the two shapes~\cite{stasseRealtimeSelfCollision2008} and can be done by finding a plane separating~$\Aone$ from~$\Atwo$.
In the rest of the paper, we focus on the generic problem of computing the distance between~$\Aone$ and~$\Atwo$, as it encapsulates the simpler Boolean collision check covered later in this section.
Results for the particular Boolean collision checking case are analyzed in the experimental section~\ref{sec:experiments}.

%%%%%%%%%%%%%%%%%%%%%%
%%%%%%%%%%%%%%%%%%%%%%
%% OPTIMIZATION PROBLEM ON THE MINKOWSKI DIFFERENCE
\vspace{0.2cm}
\noindent
\textbf{Recasting the distance computation problem onto the Minkowski difference.}
The first important idea of \citet{GJK88} is to recast the distance computation problem onto the Minkowski difference~$\Ddiff$ of the shapes, illustrated in Fig.~\ref{fig:shapes_mink_dist} and defined as follows:
\begin{equation}
    \begin{aligned}
    \Ddiff = \Aone - \Atwo = \{\bm x= \bm x_1 - \bm x_2 ~|~ \bm x_1 \in \Aone, \bm x_2\in \Atwo\} \subset \Cspace\, , 
    \end{aligned}
    \label{eq:minkowski_difference}
\end{equation}
%%% --> END EQUATION
where $\Cspace = \Rn$ is the so-called~\textit{collision space}. 
The shapes~$\Aone$ and~$\Atwo$ lie in the shape space, and the Minkowski difference~$\Ddiff$ lies in the collision space.
Although both the shape space and the collision space are in~$\Rn$, we make a distinction between the two to highlight the change in perspective.
We stress that the Minkowski difference~$\Ddiff$ is~\textit{specific} to shapes~$\Aone$ and~$\Atwo$.
If the relative position or relative orientation between~$\Aone$ and~$\Atwo$ changes, their Minkowski difference changes accordingly.

The following properties hold for the Minkowski difference~$\Ddiff$:
\begin{enumerate}
    \item Since~$\Aone$~and~$\Atwo$ are convex sets,~$\Ddiff$ is also convex.
    \item If~$\Aone$ and~$\Atwo$ are intersecting, the origin of~$\Cspace$, denoted as~$\bm 0_\Cspace$, lies inside the Minkowski difference~$\Ddiff$, \ie~\mbox{$\bm 0_{\Cspace} = \bm x_1 - \bm x_2$} for some $\bm x_1 \in \Aone$ and $\bm x_2 \in \Atwo$.
    \item If~$\Aone$ and~$\Atwo$ are not intersecting, the projection of $\bm 0_{\Cspace}$ onto~$\Ddiff$,~$\xstar = \text{proj}_{\Ddiff}(\bm 0_{\Cspace})$, corresponds to two vectors $\bm x_1^* \in \Aone$ and $\bm x_2^* \in \Atwo$, also called witness vectors in the computational geometry literature~\cite{ericsonRealTimeCollisionDetection}.
    Contrary to~$\xstar$, these vectors~$\bm x_1^*$ and~$\bm x_2^*$ are not necessarily unique, as is the case for non-strictly convex shapes such as two parallel boxes.
    \item Finally, we always have~$\norm{\bm{x}^*} = \dshapes$.
\end{enumerate}
This final remark allows us to recast the distance computation problem~\eqref{eq:distance_computation_problem_shapes} onto the Minkowski difference as follows:
%%% EQUATION
\begin{equation}
\begin{aligned}
    d_{1,2} &= \min_{\bm x \in \Ddiff } \, \norm{\bm x - \bm 0_{\Cspace} }^2 = \min_{\bm x \in \Ddiff } \, \norm{\bm x}^2\, .
\end{aligned}
\label{eq:distance_computation_problem_minkowski_difference}
\end{equation}
%%% --> END EQUATION
The convex optimization problem~\eqref{eq:distance_computation_problem_minkowski_difference} is equivalent to~\eqref{eq:distance_computation_problem_shapes} and is known as a Minimum-Norm Point problem in the optimization literature~\cite{bach2013learning,lacoste-julienGlobalLinearConvergence2015a,wolfeFindingNearestPoint1976a}.
In our case,~$\bm 0_{\Cspace} \in \Cspace = \Rn$ is the null vector \ie the origin of the collision space.
We thus aim at finding the point in~$\Ddiff$ with the lowest norm.
This vector~$\xstar$ is the~\textit{optimal solution} to~\eqref{eq:distance_computation_problem_minkowski_difference}, given by~\mbox{$d_{1,2} = \norm{\xstar}^2 = \dshapes^2$}.

Directly computing the Minkowski difference~$\Ddiff$ is neither analytically tractable nor computationally efficient.
Most of the first and second order methods for constrained convex optimization problems, such as projected gradient descent or interior point methods~\cite{boydConvexOptimization2004}, are thus sub-optimal choices.
However, computing support vectors of the Minkowski difference~$\Ddiff$, a notion defined hereafter in this section, is relatively simple and largely demonstrated by~\citet{GJK88}.
Solving convex optimization problems by computing support vectors is the foundation of the Frank-Wolfe algorithm and its variants~\cite{jaggi13}, as we discuss next.

%%%%%%%%%%%%%%%%%%%%%%
%%%%%%%%%%%%%%%%%%%%%%
%% FRANK-WOLFE
\vspace{0.2cm}
\noindent
\textbf{Distance computation using the Frank-Wolfe algorithm.}
The Frank-Wolfe algorithm (FW)~\cite{frankAlgorithmQuadraticProgramming1956} is one of the oldest convex optimization methods and solves the following constrained optimization problem:
%%% --> EQUATION
\begin{equation}
    \begin{aligned}
       f(\xstar) = \min_{\bm x \in \mathcal D} \, f(\bm x),
    \end{aligned}
    \label{eq:FW_optimization_problem}
\end{equation}
%%% --> END EQUATION
where \mbox{$f: \mathbb{R}^n \rightarrow \mathbb{R}$} is a convex and differentiable function and $\mathcal D$ is a compact convex set.
For our distance computation problem~\eqref{eq:distance_computation_problem_minkowski_difference}, we use~$f(\bm x) = \norm{\bm x}^2$ and the Minkowski difference~$\Ddiff$ as convex constraint set.
As a side note, the following discussed algorithms all require an initial starting point~$\bm x_0 \in \Ddiff$.
Shapes used in physics engines are usually attached to a frame in order to keep track of their position and orientation in space. 
We denote~$\bm c^1 \in \Aone$ and~$\bm c^2 \in \Atwo$ the origins of the frames attached to~$\Aone$ and~$\Atwo$, respectively.
In the rest of this paper, we take~$\bm x_0 = \bm c^1 - \bm c^2$. 
%%%%%%%%%%
%%% ALGORITHM: FULL FW WITH LINESEARCH
\begin{algorithm}[t]
    \caption{Frank-Wolfe algorithm with linesearch~\cite{jaggi13}}
    % \textbf{Let} $\bm x_0 \in \Ddiff$, $\gamma_k = \frac{k+1}{k+2} \in [0,1]$, $\epsilon >0$\\
    \textbf{Let} $\bm x_0 \in \Ddiff$, $\epsilon >0$\\
    \textbf{For} \texttt{k=0, 1, ...} \textbf{do}
    \begin{algorithmic}[1]
        \State $\bm d_k = \nabla f(\bm x_k)$ \Comment{Direction of support}
        \label{eq:FW_algo_support_direction}
        \State $\bm s_k \in \argmin_{\bm s \in \Ddiff} \langle \bm d_k, \bm s \rangle (= S_{\Ddiff}(\bm d_k))$ \Comment{Support~\eqref{eq:FW_support_function_in_direction}}
        \label{eq:FW_algo_support_call}
        \State \textbf{If~$g_{FW}(\bm x_k) \leq \epsilon \,$, return~$f(\bm x_k)$} \Comment{Duality gap~\eqref{eq:FW_duality_gap}}
        \label{eq:FW_algo_duality_gap}
        %
        % \State \texttt{$\bm x_{k+1} = \gamma_k \bm x_k + (1 - \gamma_k) \bm s_k$} \Comment{Iterate update}
        % \label{eq:FW_algo_update}
        \State $\gamma_k = \argmin_{\gamma \in [0,1]} f(\gamma \bm x_k + (1 - \gamma) \bm s_k)$ \Comment{Linesearch}
        \label{eq:FW_algo_linesearch}
        \State $\bm x_{k+1} = \gamma_k \bm x_k + (1 - \gamma_k) \bm s_k$ \Comment{Update iterate}
        \label{eq:FW_algo_update}
        \NoNumber{\textit{In the case of the distance computation problem~\eqref{eq:distance_computation_problem_minkowski_difference}, where~$f(\bm x) = \norm{\bm x}^2$, line~\ref{eq:FW_algo_linesearch}-\ref{eq:FW_algo_update} correspond to projecting~$\bm 0_{\Cspace}$ on the segment~$[\bm x_k, \bm s_k]$}:}
        \State $\bm x_{k+1} = \text{proj}_{[\bm x_k, \bm s_k]}(\bm 0_{\Cspace})$ \Comment{Project~$\bm 0_{\Cspace}$ on~$[\bm x_k, \bm s_k]$}
%         \label{eq:FW_algo_projection}
    \end{algorithmic}
    \label{alg:FW_vanilla_linesearch}
\end{algorithm}
%%% --> END ALGORITHM
%%%%%%%%%%
%%%%%%%%%%

The FW algorithm, summarized in Alg.~\ref{alg:FW_vanilla_linesearch}, is a gradient-descent method. 
It consists in iteratively applying two steps in order to converge towards the optimal solution $\xstar$ of~\eqref{eq:FW_optimization_problem}.
If we denote by $\bm x_k$ the estimate of $\xstar$  at iteration $k$, these two steps correspond to:
\begin{enumerate}
    \item First, we compute a support vector~$\bm s_k$ in the direction of~$\nabla f(\bm x_k)$, by solving a linear optimization problem on~$\Ddiff$.
    \item Second, we update our current iterate~$\bm x_k$ to obtain~$\bm x_{k+1}$, by taking a convex combination of the current iterate~$\bm x_k$ and the computed support vector~$\bm s_k$.
\end{enumerate}

%%% FIGURE
\begin{figure}[!t]
    \centering
    \includegraphics[width=0.98\linewidth]{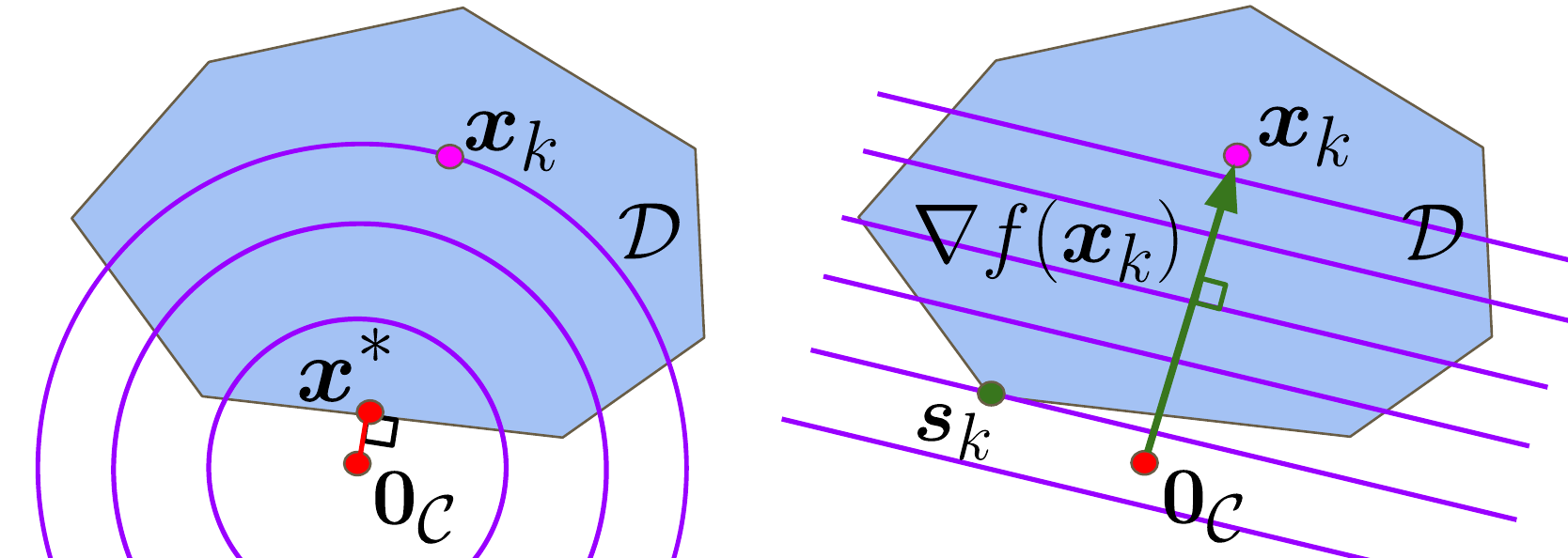}
    \caption{\small\textbf{Computing a support vector~$\bm s_k$ in direction~$\nabla f(\bm x_k)$ on convex set~$\Ddiff$.}
        We illustrate with the example of distance computation.
        On the left, we draw the Minkowski difference~$\Ddiff$ which point of minimum norm (MNP) is~$\xstar$ \ie~$\xstar$ is the projection of~$\bm 0_{\Cspace}$ onto~$\Ddiff$,~$\xstar = \text{proj}_{\Ddiff}(\bm 0_{\Cspace})$. The iterate at iteration~$k$ of the FW algorithm is~$\bm x_k$.
        In purple we draw the level sets of the function~$f(\bm x) = \norm{\bm x}^2$.
        On the right, we draw in purple the level sets of the linearization of~$f$ at iterate~$\bm x_k$,~$h_k$.
        The first step of the FW algorithm is to compute support vector~$\bm s_k$ in the direction of~$\nabla f(\bm x_k)$ (green arrow),~$\bm s_k \in S_{\Ddiff}(\nabla f(\bm x_k))$.
        In the second step of the FW algorithm, we compute~$\bm x_{k+1}$ as a convex combination of~$\bm x_k$ and~$\bm s_k$ \ie~$\bm x_{k+1}$ is a point on the segment~$[\bm x_k, \bm s_k]$.
    }
    \label{fig:level_sets_and_support_vector}
\end{figure}
%%% --> END FIGURE

In the following we detail these steps in the context of distance computation.
At iteration~$k$, the current iterate~$\bm x_k$ is the estimate of the optimal solution~$\xstar$ and~$f(\bm x_k)$ is the estimate of the optimal value of~\eqref{eq:FW_optimization_problem},~$f(\xstar)$, at iteration~$k$.
We write the linearization of the function $f$ at $\bm x_k$ and denote it as~$h_k$:
%%% --> EQUATION
\begin{equation}
    \begin{aligned}
    %&h_k: \mathbb{R}^n \rightarrow \mathbb{R}\\
    &h_k(\bm s) = f(\bm x_k) + \langle \nabla  f(\bm x_k), \bm s - \bm x_k \rangle
    \end{aligned}
    \label{eq:FW_linearization_of_function}
\end{equation}
%%% --> END EQUATION
where~$\bm s$ is a vector of~$\Rn$, ~$\nabla f(\bm x_k)$ is the gradient of~$f$ at~$\bm x_k$ and~$\langle \cdot, \cdot \rangle$ denotes the dot product between two vectors of~$\Rn$.

\textbf{$\hookrightarrow$ Step 1.}
The first step of the FW algorithm at iteration~$k$ consists of finding a minimizer~$\bm s_k \in \Ddiff$ of $h_k$ on the convex set~$\Ddiff$ (line~\ref{eq:FW_algo_support_call} in Alg.~\ref{alg:FW_vanilla_linesearch}).
Such a vector~$\bm s_k$ is called a \textit{support vector} of~$\Ddiff$ or simply a~\textit{support} and is defined as follows:
%%% --> EQUATION
\begin{equation}
    \begin{aligned}
        \bm s_k &\in \argmin_{\bm s \in \Ddiff} h_k(\bm s) = \argmin_{\bm s \in \Ddiff} \langle \nabla f(\bm x_k), \bm s\rangle \, .\\
    \end{aligned}
    \label{eq:FW_support_vector}
\end{equation}
%%% --> END EQUATION
Fig.~\ref{fig:level_sets_and_support_vector} gives a graphical understanding of support~$\bm s_k$.
The vector~$\bm s_k$ belongs to~$\Ddiff$ and is in the \textit{most} opposite direction w.r.t.~$\nabla f(\bm x_k)$.
In order to highlight the importance of the direction in which a support~$\bm s_k$ is computed, we now introduce the notion of~\textit{support direction} and~\textit{support function}. 
Given a support direction~$\bm d \in \Rn$, the support function~$S_{\Ddiff}$ returns a set of~$\Ddiff$ and is defined as:
%%% --> EQUATION
\begin{equation}
    \begin{aligned}
    &S_{\Ddiff}(\bm d) = \argmin_{\bm s \in \Ddiff} \langle \bm d, \bm s \rangle \subset \Ddiff \, .
    \end{aligned}
    \label{eq:FW_support_function_in_direction}
\end{equation}
%%% --> END EQUATION
The support function~$S_{\Ddiff}$ may return a set with more than one vector.
We only need to use one vector of this set.
Thinking in terms of direction of support allows us to understand that this direction can be rescaled while preserving the output of the support function:
%%% --> EQUATION
\begin{equation}
    \begin{aligned}
    \forall \bm d \in \Rn, \, \forall \alpha>0, \, S_{\Ddiff}(\alpha \bm d) = S_{\Ddiff}(\bm d).
\end{aligned}
    \label{eq:FW_support_funtion_direction_scaling}
\end{equation}
%%% --> END EQUATION
A support~$\bm s_k \in \Ddiff$ at iteration~$k$ is thus computed in the direction~$\bm d_k = \nabla f(\bm x_k)$ and belongs to~$S_{\Ddiff}(\nabla f(\bm x_k))$,~$\bm s_k \in S_{\Ddiff}(\nabla f(\bm x_k))$.

\begin{figure}[!t]
    \centering
    \includegraphics[width=0.98\linewidth]{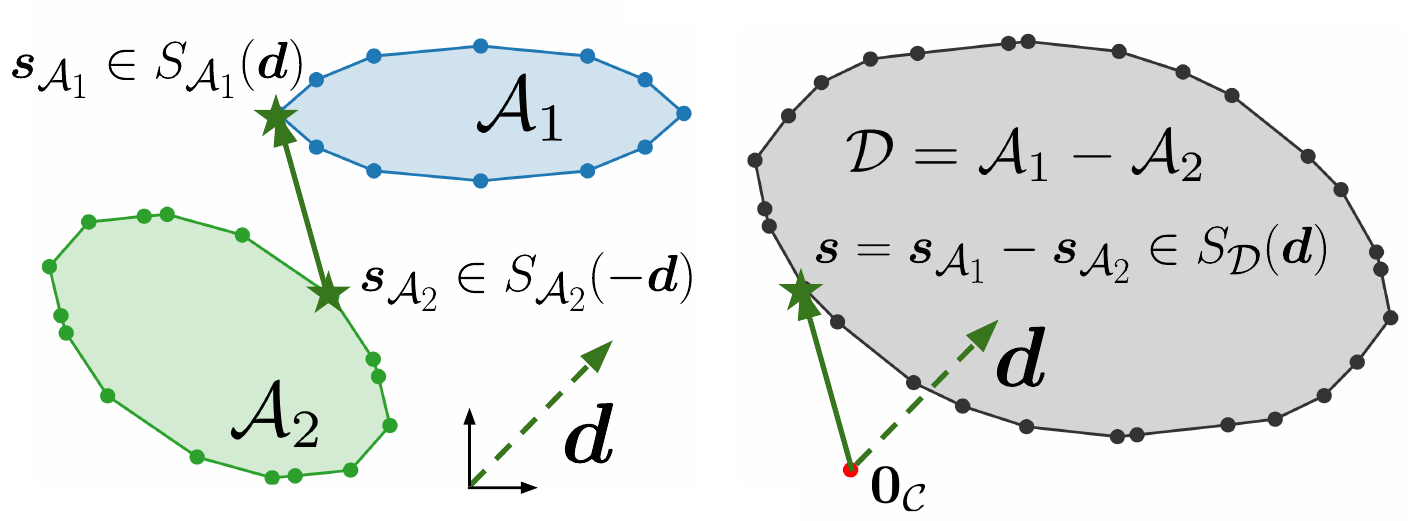}
    \caption{\small\textbf{Computing a support vector on the Minkowski difference using support vectors on the individual shapes}. 
        The vector~$\bm s_{\Aone}$ is a support vector of shape~$\Aone$ in direction~$\bm d$.
        The vector~$\bm s_{\Atwo}$ is a support vector of shape~$\Atwo$ in direction~$-\bm d$.
        The constructed vector~$\bm s = \bm s_{\Aone} - \bm s_{\Atwo}$ is a support vector of the Minkowski difference~$\Ddiff$ in the direction~$\Ddiff$.
    }
    \label{fig:minkowski_difference_linearity_support}
    \vspace{-0.5cm}
\end{figure}

We now explain how to compute the support vector~$\bm s_k$ in the case of the distance computation problem~\eqref{eq:distance_computation_problem_minkowski_difference} where we minimize~$f(\bm x) = \norm{\bm x}^2$ on the Minkowski difference~$\Ddiff$ of~$\Aone$ and~$\Atwo$.
First, we have~$\nabla f(\bm x) = 2 \bm x$.
Therefore, in the case of problem~\eqref{eq:distance_computation_problem_minkowski_difference}, it follows that:
%%% --> EQUATION
\begin{equation}
    \begin{aligned}
        \bm s_k \in S_{\Ddiff}(\bm x_k) = \argmin_{\bm s \in \Ddiff} \langle \bm x_k, \bm s \rangle \, .
    \end{aligned}
    \label{eq:FW_support_funtion_in_direction_distance_problem}
\end{equation}
%%% --> END EQUATION
As demonstrated by~\citet{GJK88}, any vector $\bm s \in S_{\Ddiff}(\bm d)$ related to the Minkowski difference can be decomposed as the difference between two support vectors $\bm s_{\Aone} \in S_{\Aone}(\bm d)$ and~$\bm s_{\Atwo} \in S_{\Atwo}(-\bm d)$ over the two individual shapes, leading to the following relation:
%%% --> EQUATION
\begin{equation}
    \begin{aligned}
        \bm s = \bm s_{\Aone} - \bm s_{\Atwo} \in S_{\Ddiff}(\bm d).
    \end{aligned}
    \label{eq:minkowski_difference_linearity_support}
\end{equation}
%%% --> END EQUATION
Equation~\eqref{eq:minkowski_difference_linearity_support} shows that we can construct a support of the Minkowski difference from the supports of the original shapes.
This property highlights the powerful change of perspective of working on the Minkowski difference.
Indeed, there exists a large number of shapes for which computing supports is simple: spheres, ellipsoids, cylinders, capsules, polytopes etc.~\cite{ericsonRealTimeCollisionDetection,bergenFastRobustGJK1999,vanDenBergenCollisionDetectionBook}.
Fig.~\ref{fig:minkowski_difference_linearity_support} illustrates the construction of a support of the Minkowski difference~$\Ddiff$ using the supports of the original shapes~$\Aone$ and~$\Atwo$.
%%% FIGURE
%%% --> END FIGURE

\textbf{$\hookrightarrow$ Step 2.}
Once a support vector~$\bm s_k \in S_{\Ddiff}(\bm x_k)$ has been computed, we update the iterate~$\bm x_k$ to obtain~$\bm x_{k+1}$ by taking a convex combination between~$\bm s_k$ and~$\bm x_k$.
The original FW algorithm uses a parameter-free update:
%%% EQUATION
\begin{equation}
    \begin{aligned}
    \bm x_{k+1} = \gamma_k \bm x_k + (1 - \gamma_k) \bm s_k \, ,
    \end{aligned}
    \label{eq:FW_update}
\end{equation}
%%% --> END EQUATION
where~\mbox{$\gamma_k = \frac{k+1}{k+2} \in [0,1]$} controls the step size.
Alternatively, a linesearch can be carried out to find a better iterate~$\bm x_{k+1}$ (line~\ref{eq:FW_algo_linesearch} in Alg.~\ref{alg:FW_vanilla_linesearch}):
%%% EQUATION
\begin{equation}
    \begin{aligned}
    &\bm \gamma_k = \argmin_{\gamma \in [0,1]} f(\gamma \bm x_k + (1 - \gamma) \bm s_k)\\
    &\bm x_{k+1} = \gamma_k \bm x_k + (1 - \gamma_k) \bm s_k.
    \end{aligned}
    \label{eq:FW_update_linesearch}
\end{equation}
%%% --> END EQUATION
In the distance computation case where~$f(\bm x) = \norm{\bm x}^2$, this linesearch~\eqref{eq:FW_update_linesearch} is equivalent to projecting~$\bm 0_{\Cspace}$ onto the segment~$[\bm x_k , \bm s_k]$,~$\bm x_k = \text{proj}_{[\bm x_k, \bm s_k]}(\bm 0_{\Cspace})$ (line~\ref{eq:FW_algo_linesearch} in Alg.~\ref{alg:FW_vanilla_linesearch}).
Since~$\Ddiff$ is convex, both~\eqref{eq:FW_update} and~\eqref{eq:FW_update_linesearch} updates are guaranteed to remain in~$\Ddiff$. 

%%%%%%%%%%%%%%%%%%%%%%
%%%%%%%%%%%%%%%%%%%%%%
%%% THE DUALITY GAP
\vspace{0.2cm}
\noindent
\textbf{Stopping criteria.}
As Frank-Wolfe deals with convex problems, the~\textit{duality gap} associated to problem~\eqref{eq:FW_optimization_problem} can be used as a stopping criterion.
Due to its convexity, the function~$f$ is always above its linearization.
Otherwise said, for any~\mbox{$\bm x \in \Rn$} and any~$\bm s \in \Rn$:
%%% EQUATION
\begin{equation}
    \begin{aligned}
        f(\bm s) \ge f(\bm x) +  \langle \nabla f(\bm x), \bm s-\bm x \rangle \, .
    \end{aligned}
    \label{eq:FW_global_inequality}
\end{equation}
%%% --> END EQUATION
Reworking this inequality and applying the $\min$ operator enables us to compute the Frank-Wolfe duality gap~$g_{\text{FW}}(\bm x) \in \mathbb{R}_{+}$ which gives an upper-bound on the difference $f(\bm x) - f(\bm x^\ast)$:
%%% EQUATION
\begin{equation}
    \begin{aligned}
        f(\bm x) - f(\bm x^\ast) \leq -\min_{\bm s \in \Ddiff} \langle \nabla f(\bm x),
        \bm s - \bm x\rangle = g_{\text{FW}}(\bm x) \, .
    \end{aligned}
    \label{eq:FW_global_inequality_min}
\end{equation}
%%% --> END EQUATION
In particular, at iteration~$k$ of the FW algorithm, we have:
%%% EQUATION
\begin{equation}
    \begin{aligned}
        f(\bm x_k) - f(\xstar) \leq g_{\text{FW}}(\bm x_k) = \langle \nabla f(\bm x_k), \bm x_k - \bm s_k \rangle \, ,
    \end{aligned}
    \label{eq:FW_duality_gap}
\end{equation}
%%% --> END EQUATION
where~$\bm s_k \in S_{\Ddiff}(\nabla f(\bm x_k))$ is the support vector computed at iteration~$k$ in the direction of~$\nabla f(\bm x_k)$.
The duality-gap~$g_{\text{FW}}(\bm x_k)$ serves as a convergence criterion for the Frank-Wolfe method and is cheap to compute.
Applied to the distance computation problem~\eqref{eq:distance_computation_problem_minkowski_difference}, the duality gap at iteration~$k$,~$g_{\text{FW}}(\bm x_k)$, guarantees that:
%%% EQUATION
\begin{equation}
    \begin{aligned}
       \norm{\bm x_k}^2 - \norm{\xstar}^2 \leq g_{\text{FW}}(\bm x_k) = 2\langle \bm x_k, \bm x_k - \bm s_k \rangle.
    \end{aligned}
    \label{eq:FW_duality_gap_distance_computation}
\end{equation}
%%% --> END EQUATION
Using the triangular inequality of the Euclidian norm and the convexity of the Minkowski difference~$\Ddiff$, we can show that:
%%% EQUATION
\begin{equation}
    \begin{aligned}
        \norm{\bm x_k - \xstar}^2 \leq \norm{\bm x_k}^2 - \norm{\xstar}^2 \leq g_{\text{FW}}(\bm x_k) \, .
    \end{aligned}
    \label{eq:FW_duality_gap_distance_computation_distance_to_optimal_solution}
\end{equation}
%%% --> END EQUATION
Inequality~\eqref{eq:FW_duality_gap_distance_computation_distance_to_optimal_solution} is useful in practice as it allows to finely control the desired tolerance on the distance to the optimal solution~$\xstar$ (line~\ref{eq:FW_algo_duality_gap} in Alg.~\ref{alg:FW_vanilla_linesearch}).
Indeed, if ones wants to compute an estimate $\bm x$ of the optimal solution $\xstar$ at precision $\epsilon$, meaning that \mbox{$\|\bm x - \xstar\| \leq \sqrt{\epsilon}$}, it is sufficient to check that $g_{\text{FW}}(\bm x) \leq \epsilon$.

%%%%%%%%%%%%%%%%%%%%%%
%%%%%%%%%%%%%%%%%%%%%%
%%% BOOLEAN COLLISION CHECKING
\vspace{0.2cm}
\noindent
\textbf{Boolean collision checking.}
As mentioned earlier, the problem of distance computation encompasses the problem of collision checking. 
Indeed, in collision checking, we are only interested in finding a separating plane between~$\Aone$ and~$\Atwo$, if it exists.
This is equivalent to finding a separating plane between~$\Ddiff$ and~$\bm 0_{\Cspace}$.
For \textit{any} support direction~$\bm d$, if we have:
%%% EQUATION
\begin{equation}
    \begin{aligned}
    \langle \bm d, \bm s \rangle > 0, \, \bm s \in S_{\Ddiff}(\bm d)
    \end{aligned},
    \label{eq:support_separating_plane}
\end{equation}
%%% --> END EQUATION
then the plane supported by the vector~$\bm d$ separates~$\Ddiff$ from~$\bm 0_{\Cspace}$~\cite{bergenFastRobustGJK1999}.
This also means that, in the case where the two shapes intersect, collision checking has the same computational complexity than distance computation. 
As shown in Alg.~\ref{alg:separating_plane_condition}, we add this separating plane condition before line~\ref{eq:FW_algo_support_call_active_set} in Alg.~\ref{alg:FW_vanilla_linesearch}.

\begin{algorithm}[t]
  \caption{Boolean collision checking: separating plane condition}
  \textit{Insert after line~\ref{eq:FW_algo_support_call_active_set} in Alg.~\ref{alg:FW_vanilla_linesearch}:}
  \begin{algorithmic}[1]
    \State \textbf{If~$\langle \bm d_k , \bm s_k \rangle > 0\,$, return \texttt{False}}
    \NoNumber{\textbf{If after termination~$d_{1,2}=0$, return \texttt{True}}}
  \end{algorithmic}
  \label{alg:separating_plane_condition}
\end{algorithm}
%%% --> END ALGORITHM

%%%%%%%%%%%%%%%%%%%%%%
%%%%%%%%%%%%%%%%%%%%%%
%% FULLY CORRECTED FRANK-WOLFE AND GJK
\vspace{0.2cm}
\noindent
\textbf{The Frank-Wolfe active-set.} 
As many gradient-descent algorithms, the FW method tends to zig-zag towards the optimal solution~\cite{lacoste-julienGlobalLinearConvergence2015a}, slowing down the convergence to the optimum.
This behavior is undesired and amplified if the optimal solution~$\xstar$ lies close to the boundary of the constraint set~$\Ddiff$.
In collision detection, this corresponds to the case where the two shapes are not intersecting.
This zig-zagging behavior is due to the way that Frank-Wolfe approaches the set of active constraints~\cite{lacoste-julienGlobalLinearConvergence2015a}, also called \emph{active-set} in the optimization literature~\cite{boydConvexOptimization2004}.
In the FW setting, the active-set at iteration~$k$, denoted~$W_k = \{\bm s^0,...,\bm s^r\} \subset \Ddiff$, is the set of vectors in~$\Ddiff$ used by the algorithm to maintain a convex combination of the iterate~$\bm x_k$:
%%% EQUATION
\begin{equation}
    \begin{aligned}
    &\bm x_k = \sum_{i=0}^{r} \lambda^i \bm s^i, \, \sum_{i=0}^{r}\lambda^i=1 \, \text{with } \bm s^i \in W_k \subset \Ddiff  \text{ and } \lambda^i > 0.
    \end{aligned}
    \label{eq:active_set}
\end{equation}
%%% --> END EQUATION
In Alg.~\ref{alg:FW_vanilla_linesearch_active_set}, we rewrite the FW algorithm with linesearch (Alg.~\ref{alg:FW_vanilla_linesearch}) in order to highlight the notion of active-set: 
\begin{itemize}
    \item A iteration~$k$, the active-set is only composed of~$\bm x_k$,~$W_k = \{\bm x_k\}$.
    \item The active-set~$W_k$ is then augmented by computing a support~$\bm s_k$ (line~\ref{eq:FW_algo_support_call_active_set} in Alg.~\ref{alg:FW_vanilla_linesearch_active_set}) to obtain~$\widetilde{W}_{k+1} = \{\bm x_k, \bm s_k \}$ (line~\ref{eq:FW_algo_augment_active_set} in Alg.~\ref{alg:FW_vanilla_linesearch_active_set}).
   \item We then minimize function~$f$ on the convex-hull of~$\widetilde{W}_{k+1}$,~$\text{conv}(\widetilde{W}_{k+1})$, which is simply the segment~$[\bm x_k, \bm s_k]$. 
   For the distance computation problem~\eqref{eq:distance_computation_problem_minkowski_difference}, this linesearch operation is equivalent to projecting~$\bm 0_{\Cspace}$ onto the segment~$[\bm x_k, \bm s_k]$ (line~\ref{eq:FW_algo_linesearch_active_set} in Alg.~\ref{alg:FW_vanilla_linesearch_active_set}).
   \item Finally, the active-set is updated~$W_{k+1} = \{\bm x_{k+1}\}$ (line~\ref{eq:FW_algo_update_active_set} in Alg.~\ref{alg:FW_vanilla_linesearch_active_set}).
\end{itemize}
In practice, discarding previously computed supports when updating the active-set is inefficient and causes the zig-zagging phenomenon observed in the FW algorithm~\cite{lacoste-julienGlobalLinearConvergence2015a}.
In the optimization literature, a rich and wide variety of variants of the FW algorithm have been introduced to efficiently cope with the active-set in order to improve the convergence rate of the FW method~\cite{garberPlayingNonlinearGames2013,guelatCommentsWolfeAway1986,hollowayExtensionFrankWolfe1974,pmlr-v89-kerdreux19a,wolfeFindingNearestPoint1976a}.
However, these variants remain too generic and are not suited for the specific problem of collision detection.
Instead, we propose next to incorporate the active-set strategy used in GJK within the Frank-Wolfe setting.

\vspace{0.2cm}
\noindent
\textbf{Connection between GJK and Frank-Wolfe.} 
In the case of collision detection,~\citet{GJK88} developed an efficient strategy to handle the active-set at a minimal cost.
To represent the current estimate~$\bm x_k$ and the optimal solution~$\xstar$, GJK exploits the concept of \textit{simplexes} in $\mathbb{R}^3$.
A simplex in~$\Rn$ corresponds to a set containing~\textit{at most~$n+1$} vectors of~$\Rn$ and the \textit{rank}~$r$ of a simplex is the number of vectors it contains ($0 < r \leq n+1$).
For 3-dimensional spaces, a simplex corresponds either to a point ($r=1$), a segment ($r=2$), a triangle ($r=3$) or a tetrahedron ($r=4$).
Similarly to the simplex methods for Linear Programming~\cite{dantzig2016linear}, the Carathéodory theorem~\cite{carathodoryBerVariabilitTsbereich1907} motivates the use of simplexes.
Let~$\mathcal{Y}$ be a set of~$N \geq n$ vectors in~$\Rn$,~$\mathcal{Y}=\{\bm y^i \in \mathbb{R}^n \}_{0 \leq i \leq N}$.
The Carathéodory theorem states that any vector~$\bm x \in \text{conv}(\mathcal{Y})$ can be expressed as the convex combination of at most~$n+1$ vectors of~$\mathcal Y$:
%%% EQUATION
\begin{equation}
    \bm x = \sum_{j=0}^{r} \lambda^j \bm y^j, \text{ with } \bm y^j \in \mathcal{Y}, \, \lambda^j > 0, \, \sum_{i=0}^{r} \lambda^j = 1.
    \label{eq:strict_convex_combination}
\end{equation}
%%% --> END EQUATION
Hence, any vector in~$\Ddiff$, and particularly the optimal solution~$\xstar \in \Ddiff = \text{conv}(\Ddiff)$ of the distance computation problem~\eqref{eq:distance_computation_problem_minkowski_difference}, can be identified as a convex combination of the vectors composing a simplex~$W$.
Relying on simplexes is attractive as there is no need to run any algorithm to compute the convex-hull of a simplex as they are convex by construction.
Frank-Wolf algorithms may operate on more complex active-sets, which might become hard to tackle from a computational point of view~\cite{jaggi13,lacoste-julienGlobalLinearConvergence2015a}.
In other words, the problem of finding the optimal solution $\xstar$ can be reformulated as the problem of identifying the optimal simplex $W^*$ on which $\xstar$  can be decomposed into a convex combination. 
This is exactly the approach followed by GJK that we now detail.

At iteration~$k$ of GJK, the current iterate~$\bm x_k$ is a convex combination of the vectors composing the simplex~$W_k$ of rank~\mbox{$r_k \leq n$}. 
To update~$\bm x_k$ and~$W_k$, the following procedure is applied: 
\begin{itemize}
    \item After computing support vector~$\bm s_k$ (line~\ref{eq:FW_algo_support_call_active_set} in Alg.~\ref{alg:FW_vanilla_linesearch_active_set}), we add $\bm s_k$  to~$W_k$ to obtain~\mbox{$\widetilde{W}_{k+1} = W_k \cup \{\bm s_k\}$} (line~\ref{eq:FW_algo_augment_active_set} in Alg.~\ref{alg:FW_vanilla_linesearch_active_set}).
    The set\mbox{~$\widetilde{W}_{k+1}$} is now a simplex of rank~\mbox{$\widetilde{r}_{k+1} \leq n+1$}.
    \item We then minimize function~$f(\bm x) = \norm{\bm x}^2$ on~\mbox{$\widetilde{W}_{k+1}$} to obtain~$\bm x_{k+1}$, corresponding to projecting~$\bm 0_{\Cspace}$ onto~\mbox{$\widetilde{W}_{k+1}$}:~\mbox{$\bm x_{k+1} = \text{proj}_{\text{conv}(\widetilde{W}_{k+1})}(\bm 0_{\Cspace})$}\footnote{
The efficient projection onto simplexes in~$\mathbb{R}^3$, named the \textit{distance sub-algorithm} by~\citet{GJK88}, is thoroughly covered in~\cite{ericsonRealTimeCollisionDetection,bergenFastRobustGJK1999} and its robustness is improved in~\cite{montanariImprovingGJKAlgorithm2017}.}
    (line~\ref{eq:FW_algo_linesearch_active_set} in Alg.~\ref{alg:FW_vanilla_linesearch_active_set}).
    \item We then have two cases, summarized in Alg~\ref{alg:FW_fully_corrective_simplex_GJK}:
    \begin{enumerate}[label=$\hookrightarrow$]
        \item If $\bm x_{k+1} = \bm 0_{\Cspace}$, the algorithm is stopped.
         Thus, we have~$\xstar = \bm 0_{\Cspace}$ and~$d_{1,2} = 0$ in~\eqref{eq:distance_computation_problem_minkowski_difference} (line~\ref{eq:GJK_origin_inside} in Alg.~\ref{alg:FW_fully_corrective_simplex_GJK}).
         \item Otherwise, we construct~$W_{k+1}$ from~\mbox{$\widetilde{W}_{k+1}$}.
        To do so, we retain only the minimal number of vectors in~\mbox{$\widetilde{W}_{k+1}$} needed to express~$\bm x_{k+1}$ as a convex combination (line~\ref{eq:GJK_active_set_update} in Alg.~\ref{alg:FW_fully_corrective_simplex_GJK}).
        Indeed, as $\bm 0_{\Cspace} \notin~\widetilde{W}_{k+1}$, the projection $\bm x_{k+1}$ of $\bm 0_{\Cspace}$ on $\widetilde{W}_{k+1}$ necessarily lies on a face of $\widetilde{W}_{k+1}$, and can be expressed as a convex combination of the vectors composing this face.
        This ensures that~$W_{k+1}$ is necessarily of rank~$r_{k+1} < \widetilde{r}_{k+1} \leq n + 1$.
    \end{enumerate}
\end{itemize}

Through this discussion, it is clear that GJK is a particular case of Frank-Wolfe.
More specifically, it is a sub-case of the fully-corrective Frank-Wolfe algorithm analyzed by~\citet{lacoste-julienGlobalLinearConvergence2015a}.
The strategy used by GJK to handle the active-set has proved to be very efficient in practice and renders the GJK algorithm state of the art for collision detection.
In the next section, we propose to leverage the formulation of collision detection as a Frank-Wolfe sub-case to accelerate its convergence following the well-established Nesterov acceleration paradigm~\cite{nesterov1983AMF}.

%%% ALGORITHM:
\begin{algorithm}[t]
  \caption{Frank-Wolfe algorithm with line-search (see Alg.~\ref{alg:FW_vanilla_linesearch}) rewritten with active-sets and applied to the distance computation problem~\eqref{eq:distance_computation_problem_minkowski_difference}}
  \textbf{Let~$\bm x_0 \in \Ddiff$,~$W_0 = \{\bm x_0\},~\epsilon > 0$}\\ 
  \textbf{For} \texttt{k=0, 1, ...} \textbf{do}
  \begin{algorithmic}[1]
        \State $\bm d_k = \bm x_k$ \Comment{Direction of support}
        \label{eq:FW_algo_support_direction_active_set}
        \State $\bm s_k \in S_{\Ddiff}(\bm d_k)$ \Comment{Support~\eqref{eq:FW_support_function_in_direction}}
        \label{eq:FW_algo_support_call_active_set}
        \State \textbf{If~$g_{FW}(\bm x_k) \leq \epsilon \,$, return~$f(\bm x_k)$} \Comment{Duality gap~\eqref{eq:FW_duality_gap}}
        \label{eq:FW_algo_duality_gap_active_set}
        %
        % \NoNumber{\textit{So far, we have~$W_k = \{\bm x_k\}$.}}
        %
        \State $\widetilde{W}_{k+1} = W_k \cup \{\bm s_k\}$ \Comment{Augment active-set}
        \label{eq:FW_algo_augment_active_set}
        \State $\bm x_{k+1} = \text{proj}_{\text{conv}(\widetilde{W}_{k+1})}(\bm 0_{\Cspace})$ \Comment{Project~$\bm 0_{\Cspace}$ on~$\text{conv}(\widetilde{W}_{k+1})$}
        \label{eq:FW_algo_linesearch_active_set}
        \State $W_{k+1} = \{ \bm x_{k+1} \}$\Comment{Update active-set}
        \label{eq:FW_algo_update_active_set}
  \end{algorithmic}
  \label{alg:FW_vanilla_linesearch_active_set}
\end{algorithm}
%%% --> END ALGORITHM
%%%%%%%%%%%%%%%%
%%%%%%%%%%%%%%%%
%%% ALGORITHM: FCFW/GJK
\begin{algorithm}[t]
    \caption{Fully-corrective FW using simplexes, applied to the distance computation problem~\eqref{eq:distance_computation_problem_minkowski_difference}. This algorithm is identical to GJK~\cite{GJK88}}
    \textit{In Alg.~\ref{alg:FW_vanilla_linesearch_active_set}, let~$W_0 = \emptyset$ and replace line \ref{eq:FW_algo_update_active_set} by:}
    \begin{algorithmic}[1]
      \State \textbf{If}~$\bm x_{k+1} = \bm 0_{\Cspace}$, \textbf{return}~$0$
      \label{eq:GJK_origin_inside}
      \NoNumber{\textit{If the algorithm has not terminated, update~$\widetilde{W}_{k+1}$ to retain only the smallest number of vectors needed to express~$\bm x_{k+1}$:}}
      \State $W_{k+1}=\{\bm s^1,...,\bm s^{r}\}$ where~$\bm s^1,...,\bm s^{r}$ are the smallest number of vectors in~$\widetilde{W}_{k+1}$ such that~$\bm x_{k+1}$ is a convex combination of~$\bm s^1,...,\bm s^{r}$. 
      \label{eq:GJK_active_set_update}
    \end{algorithmic}
    \label{alg:FW_fully_corrective_simplex_GJK}
\end{algorithm}
%%% --> END ALGORITHM

% --> END SECTION II

% -- SECTION III
\section{Accelerating Collision Detection}
\label{sec:accelerated_collision_detection}
Gradient descent (GD) is the backbone of many convex optimization methods and relies solely on the gradient of the objective function.
Second order methods~\cite{boydConvexOptimization2004} such as Newton methods have faster convergence rates than GD at the price of requiring the computation and the inversion of the Hessian quantities.
Momentum methods have thus been introduced in the optimization literature to provide gradient-based methods with improved convergence rates without requiring costly Hessian evaluation.
In this section we use recent work linking the \textit{Nesterov acceleration} of GD to the FW algorithm to globally accelerate collision detection.
This global acceleration of collision detection is experimentally evaluated in Sec.~\ref{sec:experiments} on several benchmarks.

%%%%%%%%%%%%
%%%%%%%%%%%%
%%% NESTEROV FOR UNCONSTRAINED OPTIMIZATION
\vspace{0.2cm}
\noindent
\textbf{Nesterov acceleration for unconstrained optimization.}
We initially consider the following \emph{unconstrained} minimization problem:
%%% --> EQUATION
\begin{equation}
    \begin{aligned}
       f(\bm x^\ast) = \min_{\bm x \in \Rn} \, f(\bm x)
    \end{aligned},
    \label{eq:unconstrained_minimization_problem}
\end{equation}
where~\mbox{$f: \mathbb{R}^n \rightarrow \mathbb{R}$} is a convex and differentiable function.
%%% --> END EQUATION
The vanilla gradient-descent algorithm consists of following the slope of~$f$ given by its gradient~$\nabla f$.
The following scheme is applied iteratively until a given convergence criterion is met (e.g., $\| \nabla f(\bm x_k) \| < \epsilon$, with $\epsilon$ the desired precision): 
%%% --> EQUATION
\begin{equation}
    \begin{aligned}
       \bm x_{k+1} = \bm x_k + \alpha_k \nabla f(\bm x_k),
    \end{aligned}
    \label{eq:gd_update}
\end{equation}
%%% --> END EQUATION
where~$\bm x_k \in \Rn$ is the current iterate and~$\alpha_k \in \mathbb{R}$ is the gradient step.
This standard setting leads to a simple implementation with linear convergence rate~($O(1/k)$).

To go beyond this linear convergence regime, acceleration techniques have been devised in the optimization community to provide quadratic convergence rate~($O(1/k^2)$)  or more~\cite{d2021acceleration}, by relying on relatively cheap gradient evaluations.
Among these gradient-descent acceleration techniques, the Nesterov acceleration~\cite{nesterov1983AMF} is one of the better studied and most popular in practice~\cite{d2021acceleration}.
It is based on accumulating previously computed gradients in a~\textit{momentum} term~$\bm d_k$ and using this momentum~$\bm d_k$ to update the current iterate~$\bm x_k$ as:
%%% --> EQUATION
\begin{subequations}
    \begin{alignat}{2}
       &\bm y_k = \bm x_k + \delta_k \bm d_{k-1}\\
       &\bm d_k = \delta_k \bm d_{k-1} + \alpha_k \nabla f(\bm y_k)\\
       &\bm x_{k+1} = \bm x_k + \bm d_k
    \end{alignat}
    \label{eq:unconstrained_nesterov}%
\end{subequations}
%%% --> END EQUATION
where~$\delta_k \in \mathbb{R}$ is the momentum parameter and~$\bm y_k \in \Rn$ is an intermediate quantity.
The role of momentum~$\bm d_k$ is to smooth the trajectory of iterates converging towards the optimum by geometrically averaging previously computed gradients.
The~$\delta_{k}$ momentum parameter is selected to prevent damping or overshooting of the iterate trajectory when going towards the optimal solution~$\xstar$.
Note that the current iterate~$\bm x_k$ is extrapolated using the momentum term~$\bm d_k$ to compute the intermediate vector~$\bm y_k = \bm x_k + \delta_k \bm d_k$.
The gradient is then computed at the vector~$\bm y_k$.
Computing the term~$\bm y_k$ leads to an anticipatory behavior in similar spirit to extra-gradient methods~\cite{d2021acceleration}.

%%%%%%%%%%%%
%%%%%%%%%%%%
%%% FRANK-WOLFE AND NESTEROV
\vspace{0.2cm}
\noindent
\textbf{The Frank-Wolfe algorithm and Nesterov acceleration.}
Recent works of~\citet{liMomentumGuidedFrankWolfeAlgorithm2021a,liHeavyBallMomentum2021} have proposed to adapt the Nesterov acceleration to the FW setting.
We propose to leverage and adapt this FW acceleration scheme to the context of collision detection, by notably extending the FW formulation of collision detection previously developed in Sec.~\ref{sec:frank_wolfe}.
% It goes as follows.

In the original FW algorithm, the support vector at iteration~$k$,~$\bm s_k$, is computed in the direction of the gradient~$\nabla f(\bm x_k)$ (line~\ref{eq:FW_algo_support_direction} in Alg.~\ref{alg:FW_vanilla_linesearch}).
In the Nesterov acceleration of FW proposed by~\citet{liMomentumGuidedFrankWolfeAlgorithm2021a}, the direction of support for computing~$\bm s_k$
is instead defined by:
%%% --> EQUATION
\begin{subequations}
    \begin{alignat}{2}
        \bm y_k &= \delta_k \bm x_k + (1-\delta_k) \bm s_{k-1}\\
        \bm d_k &= \delta_k \bm d_{k-1} + (1 - \delta_k) \nabla f(\bm y_k)\\
        \bm s_k &= S_{\Ddiff}(\bm d_k),
    \end{alignat}
    \label{eq:FW_nesterov_support}%
\end{subequations}
%%% --> END EQUATION
where~$\bm s_{k-1}$ is the support computed at iteration~\mbox{$k-1$},~\mbox{$\delta_k = \frac{k+1}{k+3} \in [0,1]$} is the momentum parameter and~$S_{\Ddiff}$ is the support function as defined in~\eqref{eq:FW_support_function_in_direction}.
As in the Nesterov acceleration for GD,~$\bm y_k$ is an intermediary vector used to evaluate the gradient~$\nabla f(\bm y_k)$.
To ensure~$\bm y_k$ stays in~$\Ddiff$, it is a convex combination of~$\bm x_k$ and~$\bm s_{k-1}$, both vectors of~$\Ddiff$.
The direction of support is then obtained by taking a convex combination of the previous support direction~$\bm d_{k-1}$ and the gradient~$\nabla f(\bm y_k)$.
%%% ALGORITHM
\begin{algorithm}[t]
  \caption{Frank-Wolfe algorithm with line search~\cite{jaggi13}}
  \caption{Nesterov-accelerated Frank-Wolfe~\cite{liMomentumGuidedFrankWolfeAlgorithm2021a}}
    \textit{In Alg.~\ref{alg:FW_vanilla_linesearch} and Alg.~\ref{alg:FW_vanilla_linesearch_active_set}, let~$\bm d_{-1} = \bm s_{-1} = \bm x_0$,~$\delta_k = \frac{k+1}{k+3}$ and replace line~\ref{eq:FW_algo_support_direction} by:}
    % , replace lines~\ref{eq:FW_algo_support_direction} and~\ref{eq:FW_algo_support_call} by:}
  \begin{algorithmic}[1]
        \State $\bm y_k = \delta_k \bm x_k + (1-\delta_k) \bm s_{k-1}$
        \State $\bm d_k = \delta_k \bm d_{k-1} + (1 - \delta_k) \nabla f(\bm y_k)$
        % \State $\bm s_k = S_{\Ddiff}(\bm d_k)$
        %
        % \NoNumber{\textit{The support is then computed as usual in the FW algorithms,~$\bm s_k \in S_{\Ddiff}(\bm x_k)$.}}
  \end{algorithmic}
  \label{alg:FW_nesterov}
\end{algorithm}
%%% --> END ALGORITHM

\citet{liMomentumGuidedFrankWolfeAlgorithm2021a} have experimentally shown that this acceleration strategy leads to a better convergence rate of the FW algorithm when compared to the original FW algorithm.
In the following, we explain how to adapt the Nesterov acceleration of FW to collision detection.

\vspace{0.2cm}
\noindent
\textbf{Adapting Nesterov fully-corrective Frank-Wolfe to distance computation.}
Preserving GJK's simplex strategy is crucial for collision detection as it greatly speeds up the vanilla FW algorithm.
Therefore, we adapt~\eqref{eq:FW_nesterov_support} accordingly as:
%%% --> EQUATION
\begin{subequations}
    \begin{alignat}{2}
        \bm y_k &= \delta_k \bm x_k + (1-\delta_k) \bm s_{k-1} \label{eq:GJK_nesterov_a}\\
        \bm d_k &= \delta_k \bm d_{k-1} + (1 - \delta_k) \nabla f(\bm y_k) \label{eq:GJK_nesterov_b}\\
        \bm s_k &= S_{\Ddiff}(\bm d_k), \label{eq:GJK_nesterov_c}\\
        \widetilde{W}_{k+1} &= W_k \cup \{\bm s_k \}, \label{eq:GJK_nesterov_d}\\
        \bm x_{k+1} &= \text{proj}_{\text{conv}(\widetilde{W}_{k+1})}(\bm 0_{\Cspace}). \label{eq:GJK_nesterov_e}
    \end{alignat}
    \label{eq:GJK_nesterov}%
\end{subequations}
These steps are also summarized in Alg.~\ref{alg:GJK_nesterov_algo}.
%%% --> END EQUATION
The update of simplex~$W_{k+1}$ from~$\widetilde{W}_{k+1}$ is then identical to the one described in Alg.~\ref{alg:FW_fully_corrective_simplex_GJK}.
The original duality gap defined in Sec.~\ref{sec:frank_wolfe}~(Eq.~\ref{eq:FW_duality_gap}) can no longer be used as a convergence criterion.
Indeed, the following inequality:
%%% EQUATION
\begin{equation*}
    \begin{aligned}
        \norm{\bm x_k - \xstar}^2 \leq g_{\text{FW}}(\bm x_k) = 2 \langle \bm x_k, \bm x_k - \bm s_k \rangle, \, \bm s_k \in S_{\Ddiff}(\bm x_k),
    \end{aligned}
\end{equation*}
%%% --> END EQUATION
is no longer valid because the support vector~$\bm s_k$ is no longer computed in the direction of the gradient~$\nabla f(\bm x_k) = 2\bm x_k$.
Next we will show that the original stopping criterion devised in Sec.~\ref{sec:frank_wolfe} cannot be used and we need to derive a new one.

\vspace{0.2cm}
\noindent
\textbf{Stopping criterion.}
As the number of iteration~$k$ increases,~$\delta_k \underset{k \rightarrow \infty}{\rightarrow} 1$ in~\eqref{eq:GJK_nesterov}.
Therefore,~$\bm d_k$ tends to be equal to $\bm d_{k-1}$~\eqref{eq:GJK_nesterov_b} and thus~$\bm s_k = \bm s_{k-1}$~\eqref{eq:GJK_nesterov_c}.
As a consequence, augmenting~$W_k$ with~$\bm s_k$ to construct~$\widetilde{W}_{k+1}$ (see \eqref{eq:GJK_nesterov_d}) and then projecting~$\bm 0_{\Cspace}$ onto~$\widetilde{W}_{k+1}$~\eqref{eq:GJK_nesterov_e} will not result in any progress.
Therefore,~$\bm x_{k+1} = \bm x_{k}$: the algorithm reaches a fixed point and is stuck on constant support direction~$\bm d$.

%%% ALGORITHM
\begin{algorithm}[t]
  \caption{Nesterov-accelerated GJK}
  \textbf{Let~$\bm x_0 \in \Ddiff$,~$W_0 = \emptyset,~\bm d_{-1}=\bm s_{-1}=\bm x_0,~\epsilon > 0$}\\
  \textbf{For} \texttt{k=0, 1, ...} \textbf{do}
  \begin{algorithmic}[1]
        \State $\delta_k = \frac{k+1}{k+3}$\Comment{Momentum parameter value}
        \State $\bm y_k = \delta_k \bm x_k + (1-\delta_k) \bm s_{k-1}$
        \label{eq:GJK_nesterov_algo_intermediary_point} \Comment{Intermediary point~\eqref{eq:GJK_nesterov_a}}
        \State $\bm d_k = \delta_k \bm d_{k-1} + (1 - \delta_k) \nabla f(\bm y_k)$
        \label{eq:GJK_nesterov_algo_support_direction}\Comment{Support dir.~\eqref{eq:GJK_nesterov_b}}
        \State $\bm s_k \in S_{\Ddiff}(\bm d_k)$ \Comment{Support~\eqref{eq:FW_support_function_in_direction}}
        %
        % \If{$g_{\text{FW}}(\bm x_k)  \leq \epsilon$} \Comment{Fixed-point~\eqref{eq:fixed_point_condition}}
        \If{$g(\bm x_k)  \leq \epsilon$} \Comment{Fixed-point condition~\eqref{eq:optimality_criterion}}
            \State \textbf{If $\bm d_k = \bm x_k\,$, return $f(\bm x_k)$} \Comment{Algorithm terminates}
            \State $\bm s_k \in S_{\Ddiff}(\nabla f(\bm x_k))$ \Comment{Compute~$\bm s_k$ in dir.~$\nabla f(\bm x_k)$}
            \NoNumber{\textbf{Replace line~\ref{eq:GJK_nesterov_algo_support_direction} by:~$\bm d_k = \bm x_k$ until termination.}}
        \EndIf
        \label{eq:GJK_nesterov_algo_support_call}
        \State $\widetilde{W}_{k+1} = W_k \cup \{\bm s_k\}$ \Comment{Augment active-set}
        \label{eq:GJK_nesterov_algo_augment_active_set}
        \State $\bm x_{k+1} = \text{proj}_{\text{conv}(\widetilde{W}_{k+1})}(\bm 0_{\Cspace})$ \Comment{Project~$\bm 0_{\Cspace}$ on~$\text{conv}(\widetilde{W}_{k+1})$}
        \label{eq:GJK_nesterov_algo_projection}
        \State \textbf{If}~$\bm x_{k+1} = \bm 0_{\Cspace}$, \textbf{return}~$0$
        \label{eq:GJK_nesterov_origin_inside}
        \State $W_{k+1}=\{\bm s^1,...,\bm s^{r}\}$ where~$\bm s^1,...,\bm s^{r}$ are the smallest number of vectors in~$\widetilde{W}_{k+1}$ such that~$\bm x_{k+1}$ is a convex combination of~$\bm s^1,...,\bm s^{r}$. 
        \label{eq:GJK_nesterov_active_set_update}
  \end{algorithmic}
  \label{alg:GJK_nesterov_algo}
\end{algorithm}
%%% --> END ALGORITHM

In order to cope with this issue, we use the following strategy. 
Suppose~$\bm x_k \neq \bm 0_{\Cspace}$.
Since~$\bm x_k = \text{proj}_{\text{conv}({W}_{k})}(\bm 0_{\Cspace})$ we have:
%%% EQUATION
\begin{equation}
    \begin{aligned}
    \forall \bm s^i \in W_k, \, \langle \bm x_k, \bm x_k - \bm s^i \rangle = 0.
    \end{aligned}
    \label{eq:orthogonality_simplex}
\end{equation}
%%% --> END EQUATION
After computing~$\bm s_k \in S_{\Ddiff}(\bm d_k)$, if we have:
%%% EQUATION
\begin{equation}
    \begin{aligned}
    \langle \bm x_k, \bm x_k - \bm s_k \rangle \neq 0,
    % , \, \bm s_k \in S_{\Ddiff}(\bm d_k).
    \end{aligned}
    \label{eq:not_fixed_point_condition}
\end{equation}
%%% --> END EQUATION
then~$\bm s_k$ is not a linear combination of vectors in~$W_k$.
Therefore, augmenting~$W_k$ with~$\bm s_k$ to obtain~$\widetilde{W}_{k+1}$ and projecting~$\bm 0_{\Cspace}$ onto~$\text{conv}(\widetilde{W}_{k+1})$ to obtain~$\bm x_{k+1}$ will result in the algorithm progressing toward the optimum~$\xstar$.
Suppose on the contrary that:
%%% EQUATION
\begin{equation}
    \begin{aligned}
    \langle \bm x_k, \bm x_k - \bm s_k \rangle = 0,
    % , \, \bm s_k \in S_{\Ddiff}(\bm d_k).
    \end{aligned}
    \label{eq:fixed_point_condition}
\end{equation}
%%% --> END EQUATION
then~$\bm s_k$ is a linear combination of vectors in~$W_k$.
Adding~$\bm s_k$ to~$W_k$ will thus not result in any progress towards the optimum.
As a consequence, Eq.~\eqref{eq:fixed_point_condition} encompasses two cases:
\begin{itemize}
    \item If the support direction~$\bm d_k$ is aligned with~$\nabla f(\bm x_k)$, Eq.~\eqref{eq:fixed_point_condition}, corresponding to~$g_{\text{FW}}(\bm x_k)=0$, matches the termination criteria of the distance computation problem and therefore we have reached the optimum leading to~\mbox{$\bm x_k = \bm \xstar$}.
    \item Otherwise, if~$\bm d_k$ is not aligned with~$\nabla f(\bm x_k)$, the algorithm cannot stop as a null duality gap is not met.
    The algorithm thus enters a cycle where it iterates until Eq.~\eqref{eq:fixed_point_condition} does not hold.
    To cope with this undesired behavior we simply stop the Nesterov acceleration as soon as Eq.~\eqref{eq:fixed_point_condition} is met and switch back to the non-accelerated version Alg.~\ref{alg:FW_fully_corrective_simplex_GJK}.
\end{itemize}
We thus define the function~$g$ such that  for any~$\bm s_k \in \Ddiff$: 
%%% EQUATION
\begin{equation}
    \begin{aligned}
    g(\bm x_k) = 2\langle \bm x_k, \bm x_k  - \bm s_k \rangle
    \end{aligned},
    \label{eq:optimality_criterion}
\end{equation}
%%% --> END EQUATION
~$g$ is used in~Alg.~\ref{alg:GJK_nesterov_algo} as an optimality criterion (~$g\leq\epsilon$) either for stopping the Nesterov acceleration in order to continue with the vanilla GJK, or as stopping criteria qualifying an optimal solution, in which case~$g=g_{\text{FW}}$ and~\eqref{eq:FW_duality_gap_distance_computation_distance_to_optimal_solution} holds.
The entire algorithm is summarized in Alg.~\ref{alg:GJK_nesterov_algo}.

\vspace{0.2cm}
\noindent
\textbf{Nesterov acceleration for non-strictly convex shapes.}
Let us explain the effect of the Nesterov acceleration on the support direction update~\eqref{eq:GJK_nesterov_b} and distinguish between strictly convex and non-strictly convex~$\Ddiff$:
\begin{itemize}
    \item  If~$\Ddiff$ is strictly convex, any vector~$\bm s$ belonging to the surface of~$\Ddiff$ has a unique corresponding direction~$\bm d$ such that~$\bm s = S_{\Ddiff}(\bm d)$.
    Here, we stress the fact that the support function~$S_{\Ddiff}$ returns only~\textit{one} vector.
    Consequently, we have~$\bm d_{k} \neq \bm d_{k-1}$ and therefore~$\bm s_k \neq \bm s_{k-1}$.
    The fixed point condition~\eqref{eq:fixed_point_condition} is thus not met unless~$\delta_k = 1$ and Nesterov acceleration continues to be applied in Alg.~\ref{alg:GJK_nesterov_algo}.
    In practice, the algorithm runs until~$\delta_k$ gets close to~$1$ or~$\bm x_k$ gets close to~$\bm 0_{\Cspace}$.
    The condition~\eqref{eq:fixed_point_condition} is then satisfied as the algorithm starts to cycle.
    The Nesterov acceleration is thus removed and the algorithm runs until the convergence criteria is satisfied, guaranteed by the Frank-Wolfe algorithm.
    % Eq. 28 is will then becom satisfied and the switch to GJK will occur. 
    \item Otherwise, if~$\Ddiff$ is non-strictly convex, multiple support directions~$\{\bm d^1, ..., \bm d^m, ...\}$ can yield the same support vector~$\bm s \in S_{\Ddiff}(\bm d^1)=...=S_{\Ddiff}(\bm d^m)=...$ etc.
    Consequently, it is possible to have~$\bm d_{k-1} \neq \bm d_k$ and~$\bm s_k = \bm s_{k-1}$.
    Therefore, even though~$\delta_k$ is not close to~$1$, the fixed point condition~\eqref{eq:fixed_point_condition} can be verified.
    The Nesterov acceleration is stopped, possibly prematurely.
\end{itemize}
The latter case is especially problematic when shapes~$\Aone$ and~$\Atwo$ are in close-proximity, which is ultimately the type of collision problems which are commonly encountered in simulation or motion planning with contacts.
In~\eqref{eq:GJK_nesterov_b}, this is due to the norm of~$\bm d_{k-1}$ being greatly predominant over the norm of~$\bm y_k$ as~$k$ increases,~\mbox{$\norm{\bm d_{k-1}} \gg \norm{\nabla f(\bm y_k)}$}.
To prevent this phenomenon on non-strictly convex~$\Ddiff$, we propose to replace~\eqref{eq:GJK_nesterov_b} by a simple heuristic which normalizes the gradient and momentum directions as follows:
%%% EQUATION
\begin{equation}
    \begin{aligned}
        \bm d_k = \delta_k \frac{\bm d_{k-1}}{\norm{\bm d_{k-1}}} + (1 - \delta_k) \frac{\nabla f(\bm y_k)}{\norm{\nabla f(\bm y_k)}}
    \end{aligned},
    \label{eq:normalize_support_direction}
\end{equation}
%%% --> END EQUATION
summarized in Alg.~\ref{alg:normalize_direction_algo}.
In Sec.~\ref{sec:experiments}, we experimentally prove this heuristic to significantly reduce the number of steps for distance computations for non-strictly convex shapes.

\begin{algorithm}[t]
  \caption{Normalize direction for non-strictly convex shapes}
  \textit{Replace line~\ref{eq:GJK_nesterov_algo_support_direction} in Alg.~\ref{alg:GJK_nesterov_algo} by:}
  \begin{algorithmic}[1]
      \State $\bm d_k = \delta_k \frac{\bm d_{k-1}}{\norm{\bm d_{k-1}}} + (1 - \delta_k) \frac{\nabla f(\bm y_k)}{\norm{\nabla f(\bm y_k)}}$
  \end{algorithmic}
  \label{alg:normalize_direction_algo}
\end{algorithm}
%%% --> END ALGORITHM

% --> END SECTION III

% SECTION IV
\section{Experiments}
\label{sec:experiments}
%%% --> FIGURE
\begin{figure*}
    \centering
    \subfloat[
        Intersecting ellipsoids.
        \label{fig:ellipsoids_cv_overlapping}
    ]{
        \includegraphics[width=0.25\linewidth]{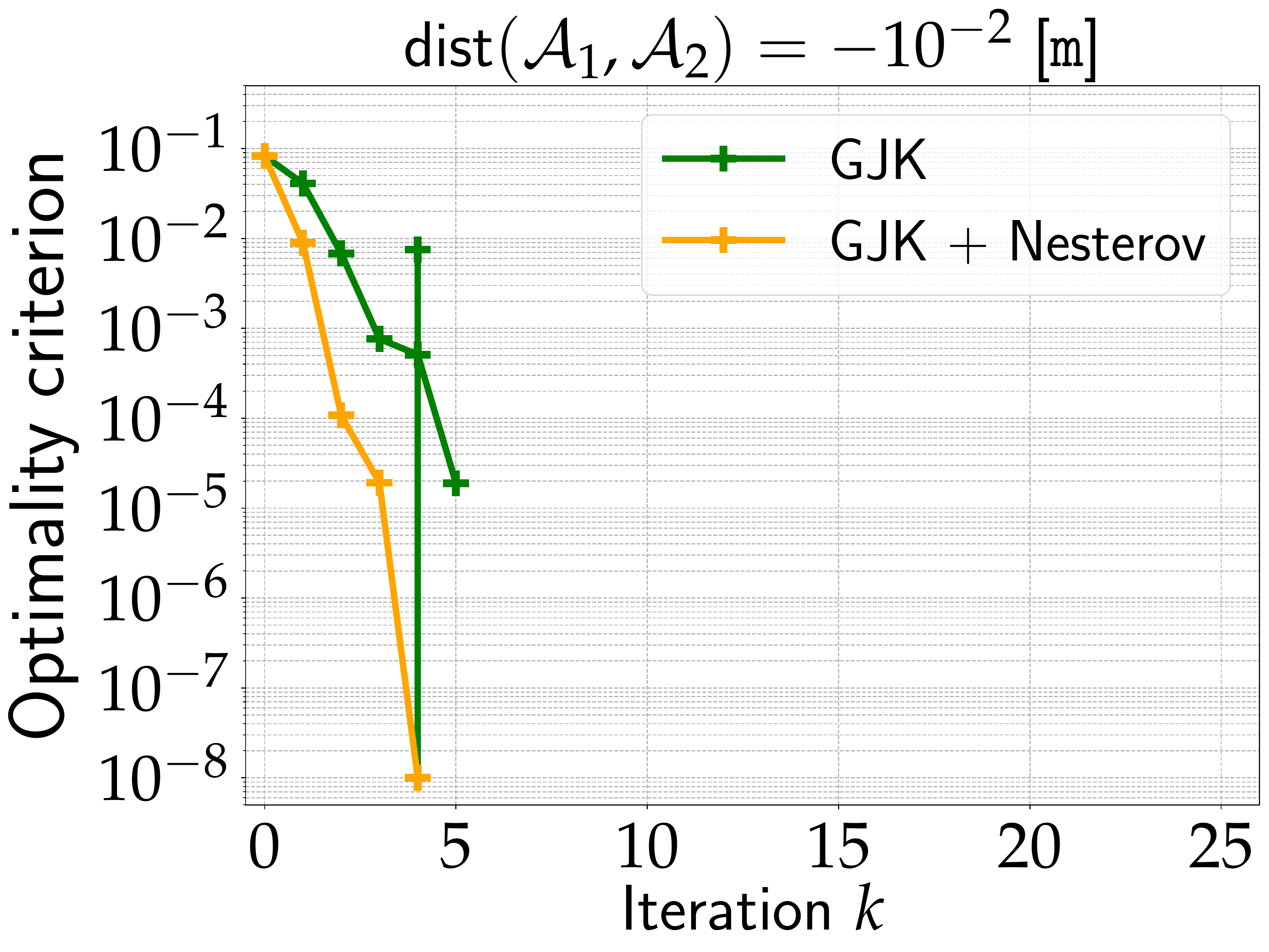}
    }\hspace{1cm}
    \subfloat[
        Close-proximity ellipsoids.
        \label{fig:ellipsoids_cv_close_proximity}
    ]{
        \includegraphics[width=0.25\linewidth]{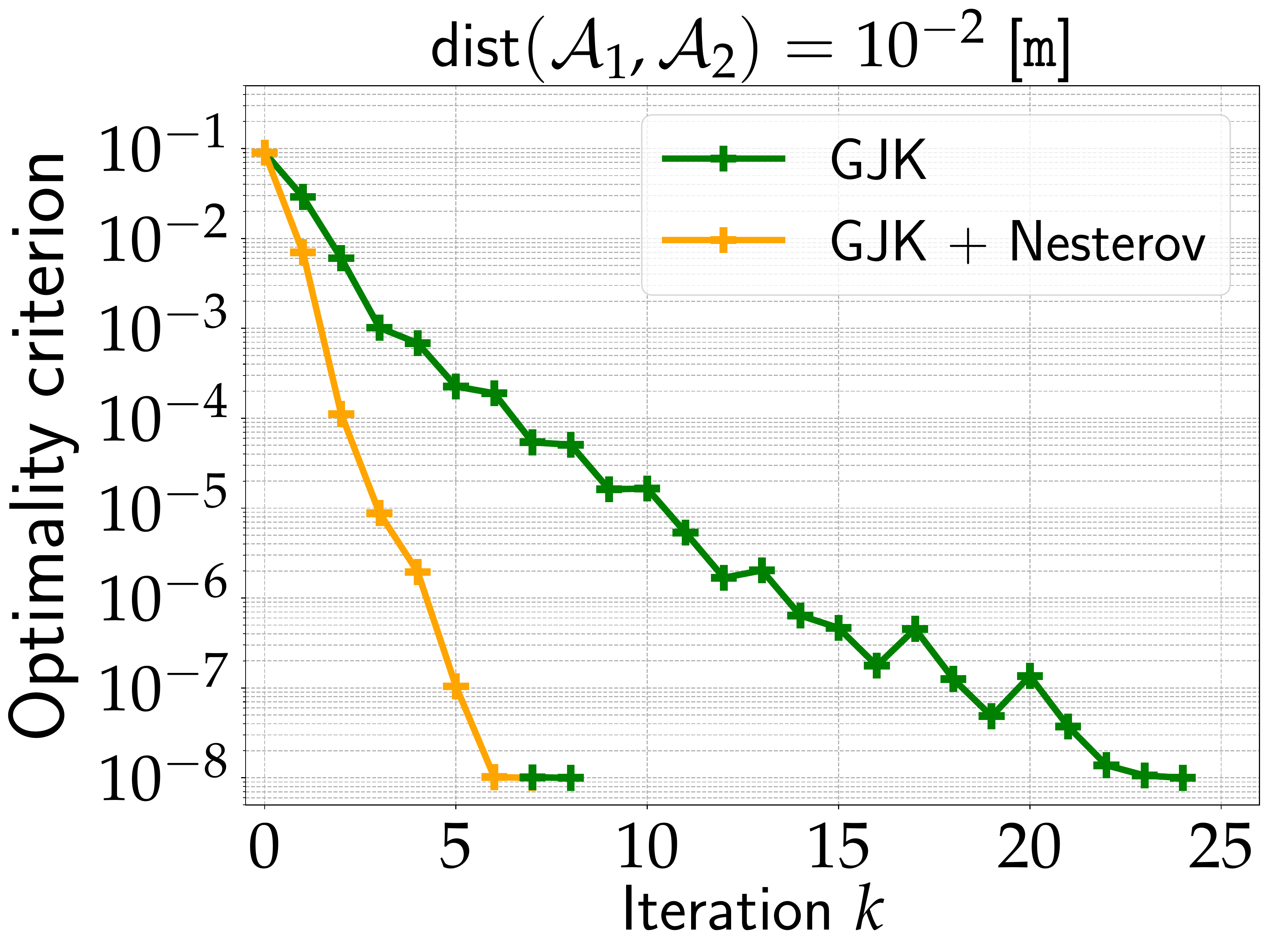}
    }\hspace{1cm}
    \subfloat[
        Distant ellipsoids.
        \label{fig:ellipsoids_cv_distant}
    ]{
        \includegraphics[width=0.25\linewidth]{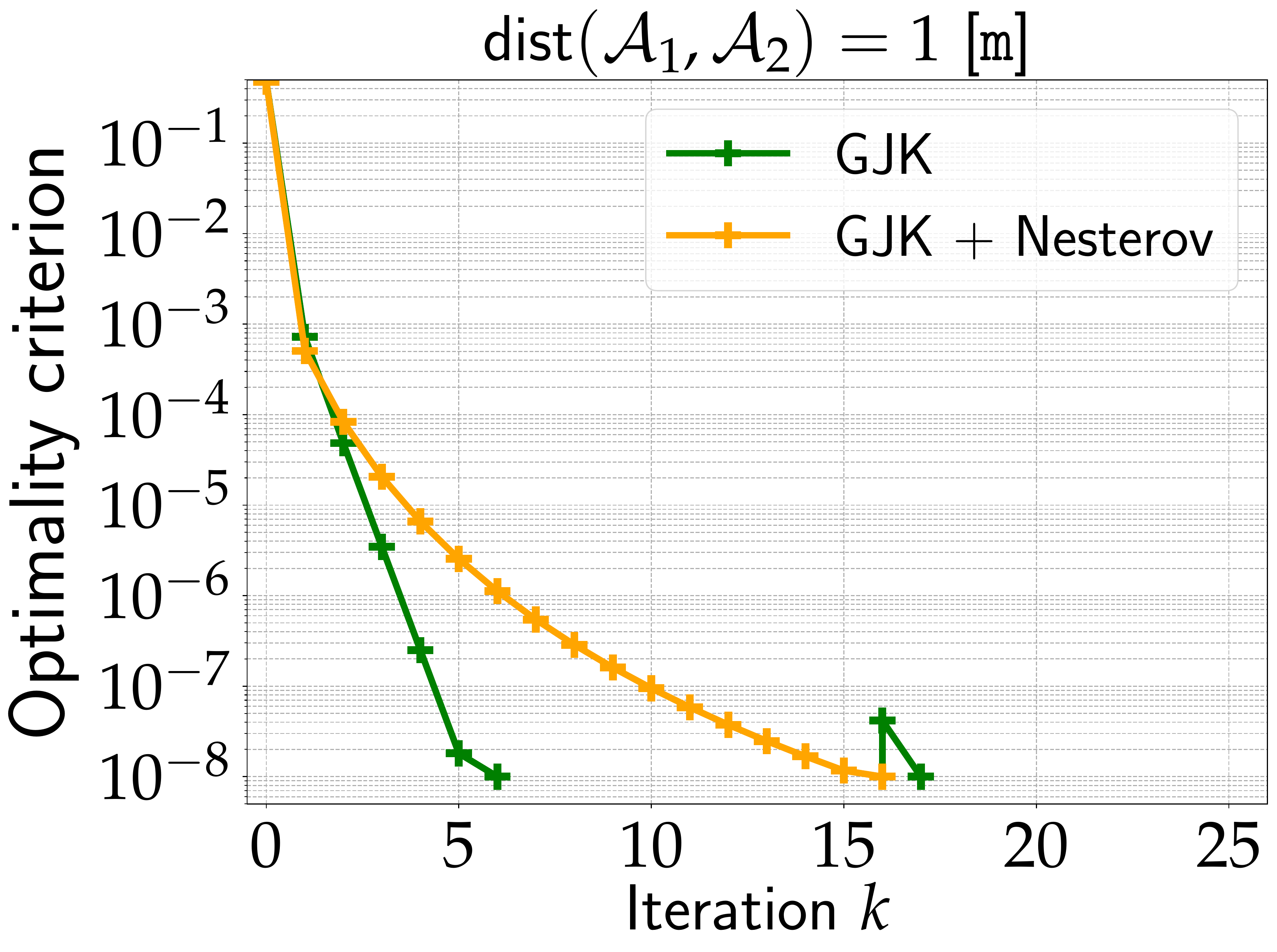}
    }
    \caption{\small
        \textbf{Convergence of GJK and Nesterov-accelerated GJK for ellipsoid collisions.}
        The $y$-axis measures the optimality criterion~\eqref{eq:optimality_criterion}.
        For Nesterov-accelerated GJK, the switch from Nesterov acceleration to vanilla GJK is denoted by the switch in color from orange to green.
        In (a), the optimality criterion defined by~\eqref{eq:optimality_criterion} is not used by any of the two methods to terminate as~$\bm 0_{\Cspace}$ lies inside~$\Ddiff$ and both algorithms terminate when~$\bm x_k = \bm 0_{\Cspace}$.
        It is however used by Nesterov-accelerated GJK to stop the Nesterov acceleration when a fixed-point is met, hence the sudden peak denoting the switch to GJK.
        Nesterov acceleration has beneficial effects in the intersecting (a) and close-proximity (b) setups which are the cases of interest, for instance, in the context of simulation and planning with contacts.
    }
    \label{fig:ellipsoids_cv}
    \vspace{-0.5cm}
\end{figure*}
%%% --> END FIGURE
%%% --> FIGURE
\begin{figure}
    \centering
    \subfloat[
        Distance computation.
        \label{fig:ellipsoids_benchmark_distance}
    ]{
        \includegraphics[width=0.48\linewidth]{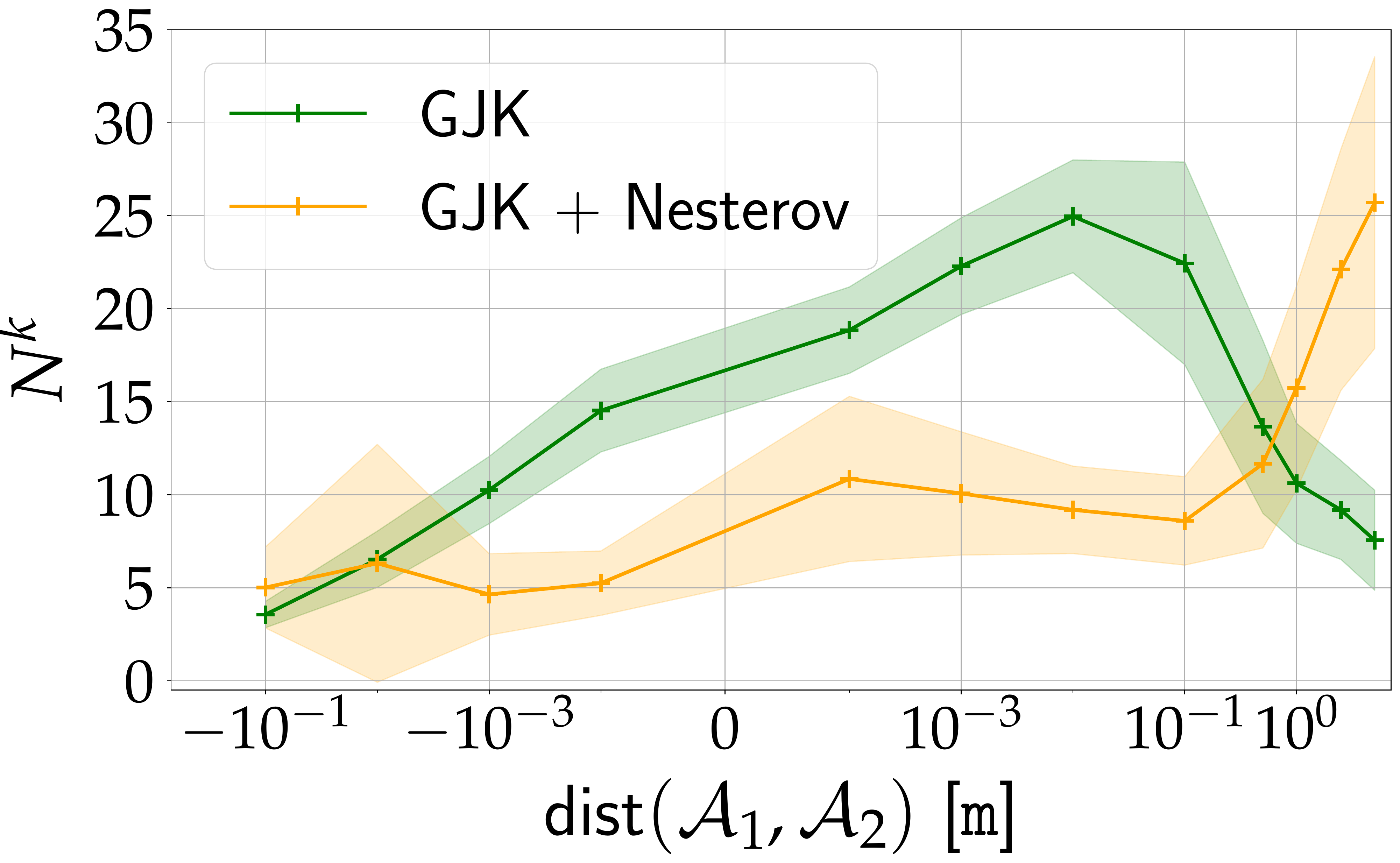}
    }
    \subfloat[
        Collision detection.
        \label{fig:ellipsoids_benchmark_collision}
    ]{
        \includegraphics[width=0.48\linewidth]{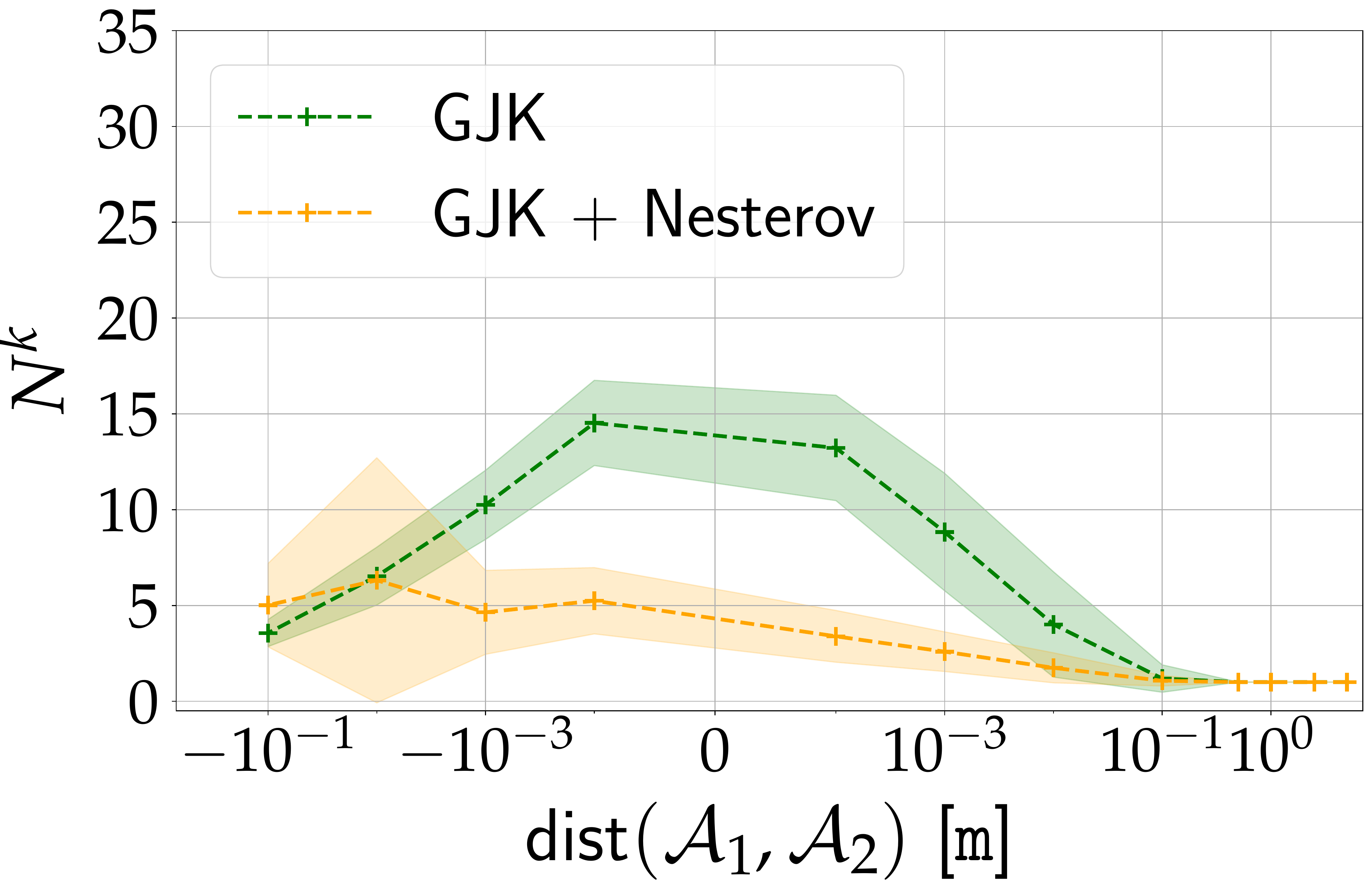}
    }
    \caption{\small
        \textbf{Comparison of Nesterov-accelerated GJK and vanilla GJK on the ellipsoid benchmark.} The graphs show the number of iterations (y-axis) vs. the signed distance between the two shapes (x-axis). The curve shows the mean value over 100,000 random trials. The shaded region corresponds to the standard deviation. The Nesterov-accelerated GJK algorithm requires fewer iterations when the shapes are in close proximity. 
    }
    \label{fig:ellipsoids_benchmark}
    \vspace{-0.5cm}
\end{figure}
%%% --> END FIGURE

In this section,  we study the performance of Nesterov-accelerated GJK (Alg.~\ref{alg:GJK_nesterov_algo}) against the vanilla GJK (Alg.~\ref{alg:FW_fully_corrective_simplex_GJK}) algorithm.
We use the \mbox{HPP-FCL} C++ library~\cite{panFCLGeneralPurpose2012,hppfclweb} and its implementation of GJK as a starting point for  the implementation of the proposed Nesterov-accelerated GJK (Alg.~\ref{alg:GJK_nesterov_algo}).
To distinguish between pairs of strictly convex and non-strictly convex shapes, we build a first benchmark only composed of pairs of ellipsoids (strictly convex shapes) and a second benchmark using pairs of standard meshes (represented by their convex hulls) which are taken from the commonly-used ShapeNet dataset~\cite{shapenet2015}.
Finally, we empirically show that the simplex strategy used by GJK and our method (discussed in Sec.~\ref{sec:frank_wolfe}) is crucial for efficient collision detection.
We show that GJK and our method significantly outperform the original FW algorithm and the recent NESMINO~\cite{nesmino2019} algorithm.
Although it differs from FW algorithms, we include the NESMINO algorithm in this analysis as its projected-gradient descent procedure is accelerated using the classic Nesterov acceleration scheme~\cite{nesterov1983AMF}.
The code to run the benchmarks is made freely available at:~\href{https://github.com/lmontaut/collision-detection-benchmark}{https://github.com/lmontaut/collision-detection-benchmark}.

\vspace{0.2cm}
\noindent
\textbf{Shape datasets.}
In the ellipsoids benchmark, the ellipsoids are randomly generated by sampling positive-definite matrices.
In total, we generate~$1000$ random pairs of ellipsoids.
Given a pair of ellipsoids, we randomly sample relative poses between the two shapes such that both objects do not intersect.
This then allows us to finely control the distance~$\dshapes$ between the objects, enabling us to measure the influence of the separation distance on the performance of the studied algorithms.
The values used for~$\dshapes$ range from~$-0.1\,\meter$ to~$1\,\meter$.
Negative values correspond to scenarii where the shapes are intersecting and~$\dshapes$ is the norm of the separating vector.
The separating vector is the vector of smallest norm needed to translate one of the two shapes such that the two shapes do not intersect.
Therefore, for each pair of ellipsoids,~$100$ random relative poses are sampled such that the shapes do not intersect.
For each relative pose, we translate the shapes along the axis given by their closest-points to study the impact of~$\dshapes$.
We then set~$\dshapes$ to fixed values between~$-0.1\,\meter$ to~$1\,\meter$.
On the other hand, the mesh dataset contains roughly~$35\text{k}$ shapes but we only use~$1.8\text{k}$ randomly selected shapes from 10 categories of the dataset to set-up the mesh benchmark.
To generate the mesh benchmark, we sample~$10\,$k random pairs of meshes and represent each shape by its convex hull.
The resulting meshes contain between~$10$ and $5000$ vertices.
About~\mbox{$50\%$} of meshes contain between~$100$ and~$1000$ vertices.
The remaining shapes are equally distributed between less than~$100$ vertices or more than~$1000$ vertices.
As in the ellipsoids benchmark,~$100$ random relative poses are sampled for each pair such that the shapes do not intersect and then set~$\dshapes$ to fixed values between~$-0.1\,\meter$ and~$1\,\meter$.
In both benchmarks, the characteristic sizes of the shapes range from a few centimeters up to a meter.
Finally, for the distance computation problem, we select a convergence tolerance of~\mbox{$\epsilon=10^{-8}$}.

\vspace{0.2cm}
\noindent
\textbf{Metrics.} 
In order to measure the performances of the Nesterov-accelerated GJK and the vanilla GJK algorithms, we measure the number of iterations~\mbox{$\numit$} to solve a given collision problem.
For the mesh benchmark, we also measure the execution time~\mbox{$\exectime$} of both methods.
To cope with CPU throttling, we solve generated collision problems~$100$ times.
We then report the average of the~\mbox{$90\%$} lowest computation times.
Since both algorithms have identical iterations which differ only in the computation of the support direction, the number of iterations is preferred to study the performance of the algorithms; the execution time serves as a reference but can vary based on the implementation and the hardware used.

%%%%%%%%%%%%%%%%%%%%%%%%%%%%%%%%%%%%%%%%%%%%%%%%%%
%%%%%%%%%%%%%%%%%%%%%%%%%%%%%%%%%%%%%%%%%%%%%%%%%%
%%%%%%%%%%%%%%%%%%%%%%%%%%%%%%%%%%%%%%%%%%%%%%%%%%
%%% ELLIPSOIDS
\subsection{Strictly convex shapes: ellipsoids}
\label{sec:ellipsoids}
\noindent
\textbf{Single collision pair.}
To qualitatively understand the effect of the Nesterov acceleration, we first study the evolution of the optimality criterion~\eqref{eq:FW_duality_gap_distance_computation_distance_to_optimal_solution} in the case of distance computation on a single pair of ellipsoids (Fig.~\ref{fig:ellipsoids_cv}).

In Fig.~\ref{fig:ellipsoids_cv_overlapping}, the shapes are intersecting and both algorithms converge to~\mbox{$\xstar = \bm 0_{\Cspace}$}.
The optimality criterion~\eqref{eq:optimality_criterion} is not used by any of the two methods to terminate.
Indeed, they stop when the projection onto the simplex~\mbox{$\widetilde{W}_k$} yields~$\bm 0_{\Cspace}$ (line~\ref{eq:GJK_nesterov_origin_inside} in Alg.~\ref{alg:GJK_nesterov_algo} and line~\ref{eq:GJK_origin_inside} in Alg.~\ref{alg:FW_fully_corrective_simplex_GJK}).
However, the criterion is still used in Nesterov-accelerated GJK to switch to GJK.
Both methods converge rapidly.
As explained in Sec.~\ref{sec:frank_wolfe}, this is expected as the optimal solution~$\xstar = \bm 0_{\Cspace}$ lies inside~$\Ddiff$.

In Fig.~\ref{fig:ellipsoids_cv_close_proximity}, the shapes are close together with a separating distance of~\mbox{$\dshapes = 0.01\,\meter$}.
Along with the previous intersecting case, this figure represents a typical scenario one could encounter in the narrow phase of the collision detection problem.
Contrary to the intersecting case, the solution~$\xstar$ now lies on the boundary of~$\Ddiff$.
The convergence of GJK is impaired due to the successive switches in support directions.
By smoothing the change of support direction, the Nesterov acceleration requires less iterations and divides by more than two the number of iterations to reach the convergence criterion compared to GJK.

When shapes are distant (Fig.~\ref{fig:ellipsoids_cv_distant}), %,~$\dshapes=1\meter$,
the problem is easier to solve for GJK.
This is due to smaller angles between successive support directions which is a consequence of the large distance between~$\bm 0_{\Cspace}$ and~$\Ddiff$.
In such a case, the Nesterov acceleration is now detrimental to the convergence of the algorithm as it induces too little successive changes in the support directions. 
This is a typical case where smoothing the gradient directions via the Nesterov acceleration does not help and takes more than triple the number of iterations for Nesterov-accelerated GJK to achieve convergence compared to vanilla GJK.
This behavior is however specific to ellipsoids and is not observed in meshes as shown below.

Fig.~\ref{fig:ellipsoids_cv} highlights three different behaviors in the convergence of both algorithms: when the shapes are~\textit{intersecting}, in \textit{close-proximity} and~\textit{distant}.
Scenarii where shapes are intersecting or in close proximity are of greater importance in general, as they correspond to contexts which are evaluated during the narrow phase.
On the opposite, distant scenarii are rejected by the broad phase, and other approximations of $\dshapes$ could be used if needed~\cite{ericsonRealTimeCollisionDetection}.

\vspace{0.2cm}
\noindent
\textbf{Statistical validation over the ellipsoids dataset.}
The convergence graphs in Fig.~\ref{fig:ellipsoids_cv} directly depend on the pair of shapes considered. 
Therefore, to get a better statistical understanding of the performance of Nesterov-accelerated GJK and vanilla GJK, we focus on the ellipsoid benchmark.

Fig.~\ref{fig:ellipsoids_benchmark_distance} and Fig.~\ref{fig:ellipsoids_benchmark_collision} show the mean and standard-deviation of the number of iterations~$\numit$ of each method for the distance computation and the Boolean collision checking problems, respectively.
In the case of distance computation (Fig.~\ref{fig:ellipsoids_benchmark_distance}), the results confirm the trend shown in the convergence graphs in Fig.~\ref{fig:ellipsoids_cv}.
When the shapes are shallowly intersecting, Nesterov-accelerated GJK converges with the same or even fewer number of iterations than vanilla GJK.
The shallower the penetration, the more Nesterov accelerates over vanilla GJK.
The irregularity in standard deviation at~$-0.01\,\meter$ is a critical zone for the momentum where the variance increases.
This could be fixed thanks to a restart strategy~\cite{d2021acceleration} and will be analysed in a future work.
When shapes are in close-proximity, the Nesterov acceleration of GJK significantly reduces  the number of iterations compared to vanilla GJK.
Finally, when shapes are distant,~\mbox{$1\,\meter \leq \dshapes$}, Nesterov-accelerated GJK is detrimental to convergence on the distance computation problem.
A similar pattern of speed-ups of the Nesterov-accelerated GJK over the vanilla GJK is shown for the collision detection problem in Fig.~\ref{fig:ellipsoids_benchmark_collision}.

%%%%%%%%%%%%%%%%%%%%%%%%%%%%%%%%%%%%%%%%%%%%%%%%%%
%%%%%%%%%%%%%%%%%%%%%%%%%%%%%%%%%%%%%%%%%%%%%%%%%%
%%%%%%%%%%%%%%%%%%%%%%%%%%%%%%%%%%%%%%%%%%%%%%%%%%
%%% SHAPENET
\subsection{Non-strictly convex shapes: meshes}
\label{sec:shapenet}

\begin{figure}
    \centering
    \subfloat[
        Distance computation.
        \label{fig:shapenet_normalization_a}
    ]{
        \includegraphics[width=0.48\linewidth]{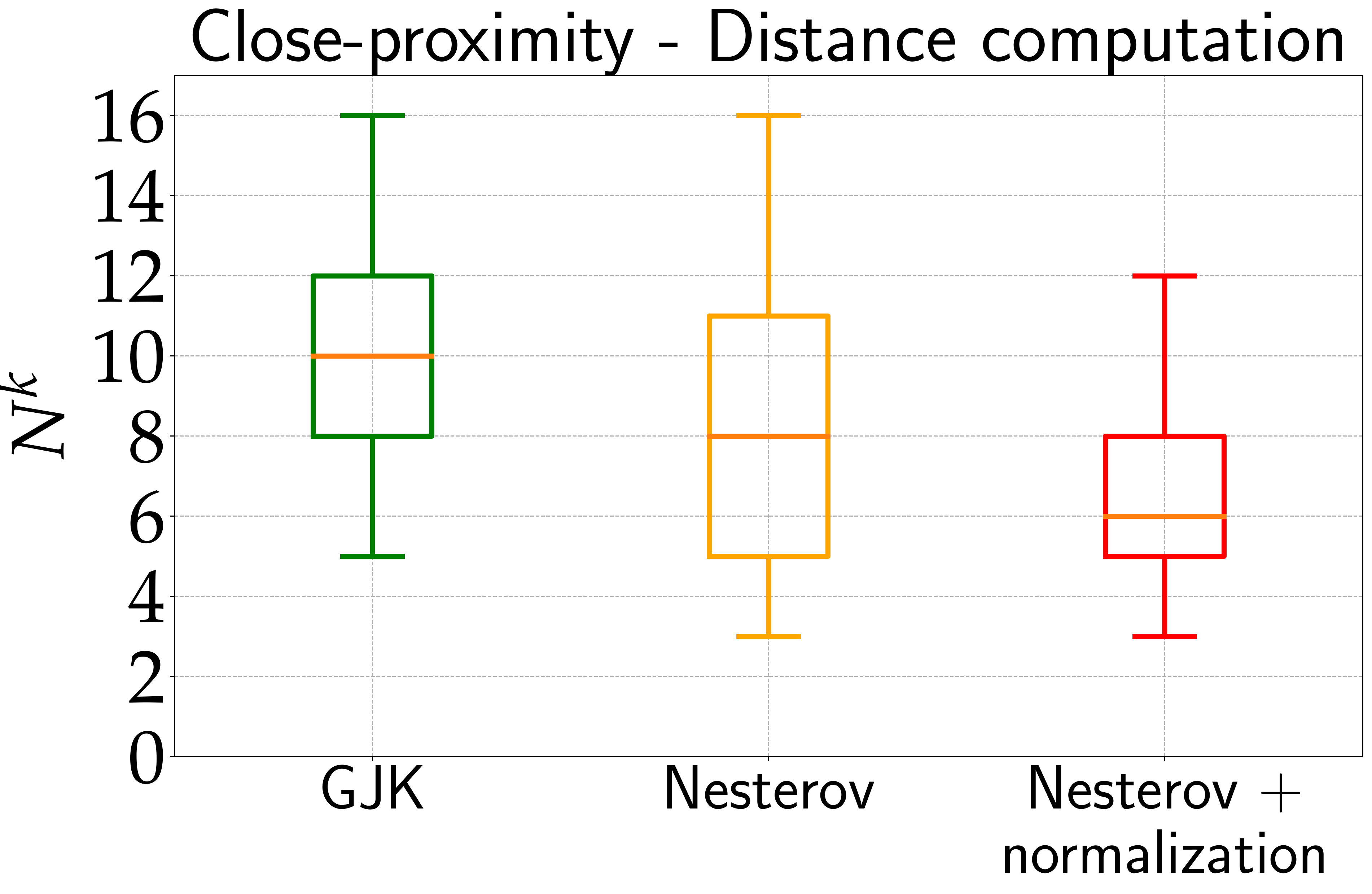}
    }
    \subfloat[
        Boolean collision check
        \label{fig:shapenet_normalization_b}
    ]{
        \includegraphics[width=0.48\linewidth]{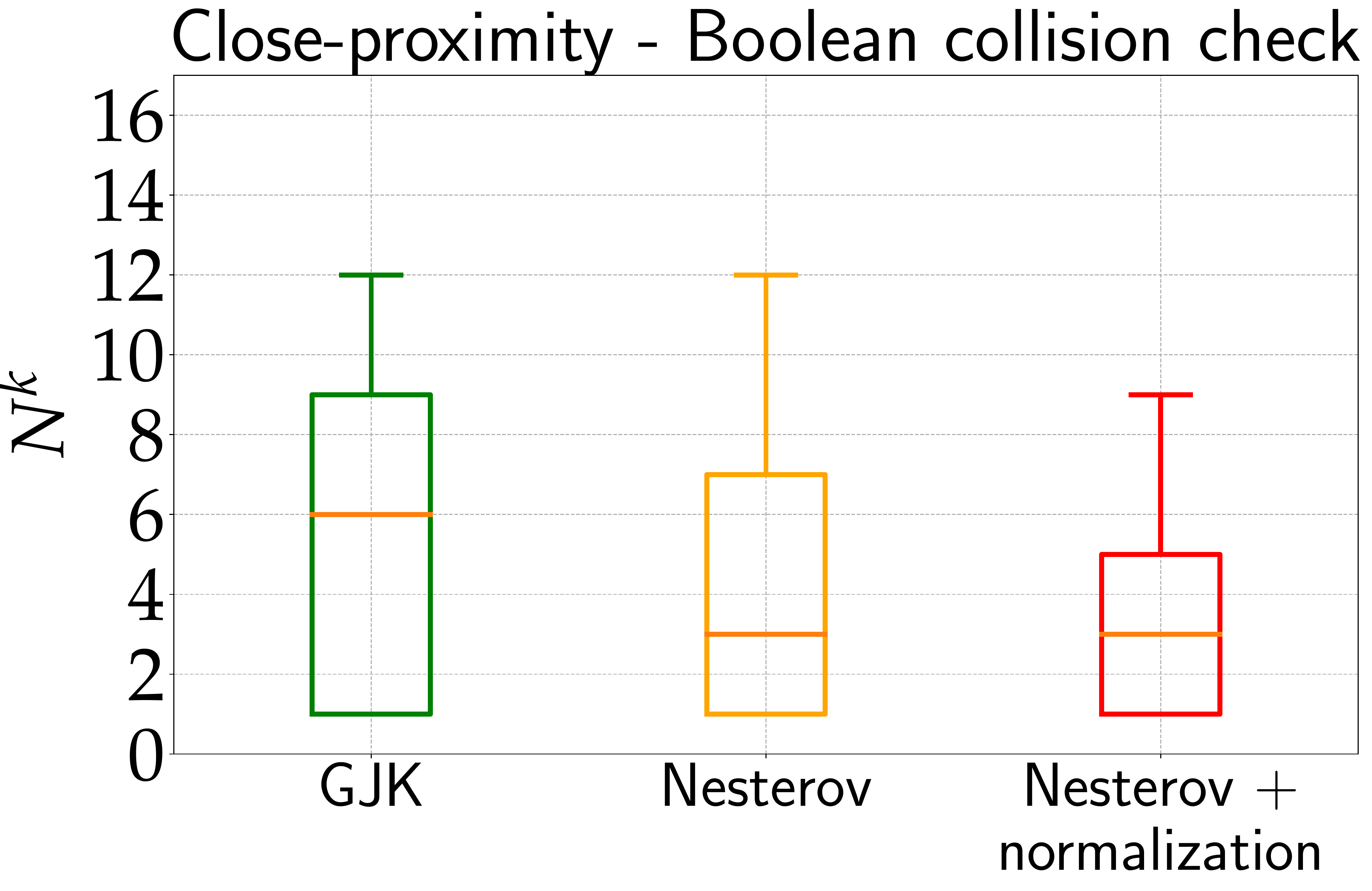}
    }
    \caption{\small
        \textbf{Importance of support direction normalization in Nesterov-accelerated GJK on the ShapeNet benchmark.} 
        The graphs show the number of iterations $\numit$ (lower is better) for GJK and Nesterov-accelerated GJK with and without support direction normalization when the two shapes are in 
        close-proximity:~\mbox{$ 0\,\meter < \dshapes \leq 0.1\,\meter$}.
        Using normalization of support direction is beneficial in Nesterov-accelerated GJK, reducing the overall number of iterations compared to GJK and non-normalized Nesterov-accelerated GJK.
    }
    \label{fig:shapenet_normalization}
    \vspace{-0.1cm}
\end{figure}

\vspace{0.1cm}
\noindent
\textbf{Effect of support direction normalization.}
For meshes, the importance of normalizing the support direction (see Eq.~\eqref{eq:normalize_support_direction}) in the Nesterov-accelerated GJK is highlighted in Fig.~\ref{fig:shapenet_normalization}.
For both the distance computation and Boolean collision checking problems, it prevents the Nesterov acceleration from reaching a fixed-point too early and consequently it reduces the overall amount of iterations needed to converge.
In the following, we thus focus only on Nesterov-accelerated GJK with support normalization and compare its performance against the vanilla GJK algorithm.
%%% --> FIGURE
%%% --> END FIGURE
%%% --> FIGURE
\begin{figure}
    \centering
    \subfloat[
        Number of iterations~$\numit$.
        Lower is better.
        \label{fig:shapenet_benchmark_distances_a}
    ]{
        \includegraphics[width=\linewidth]{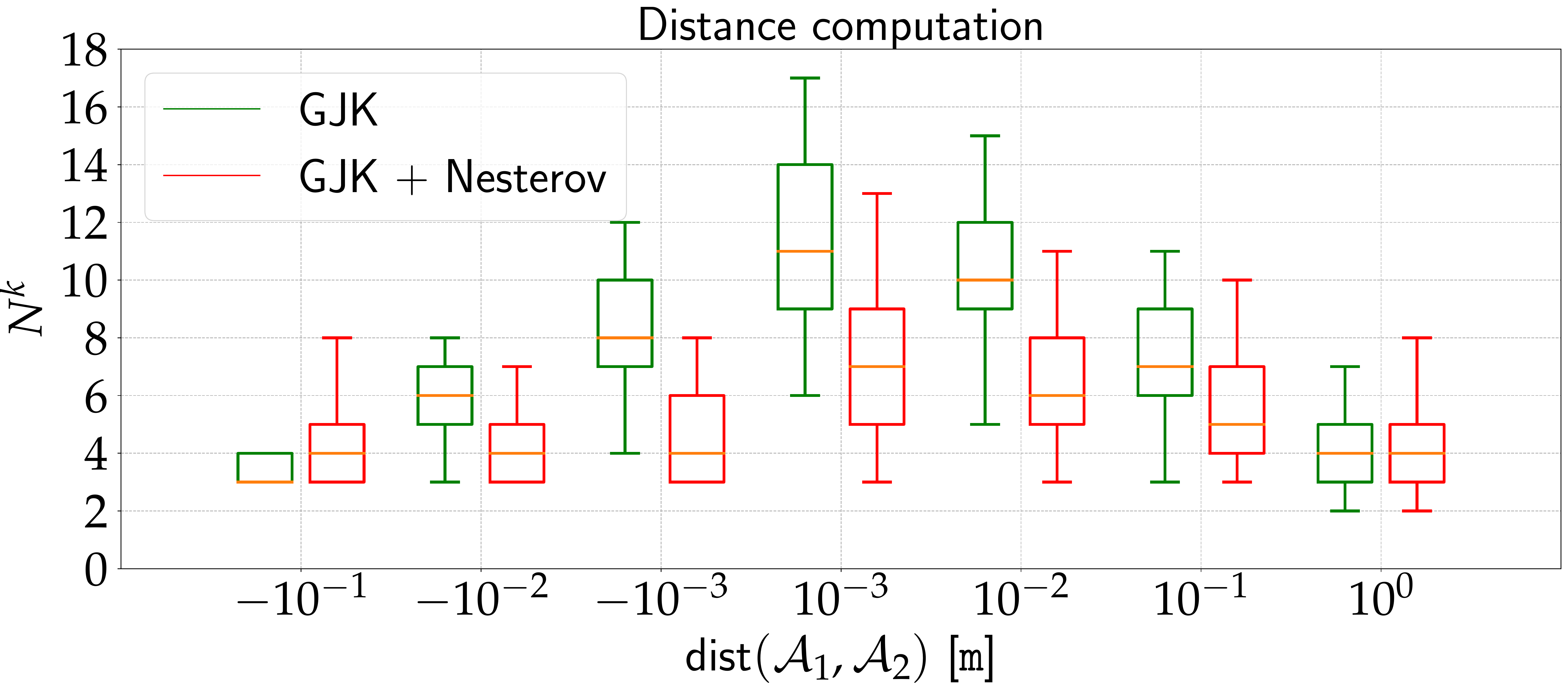}
    }\hfill\\
    \subfloat[
        Execution time~$\exectime$.
        Lower is better.
        \label{fig:shapenet_benchmark_distances_b}
    ]{
        \includegraphics[width=\linewidth]{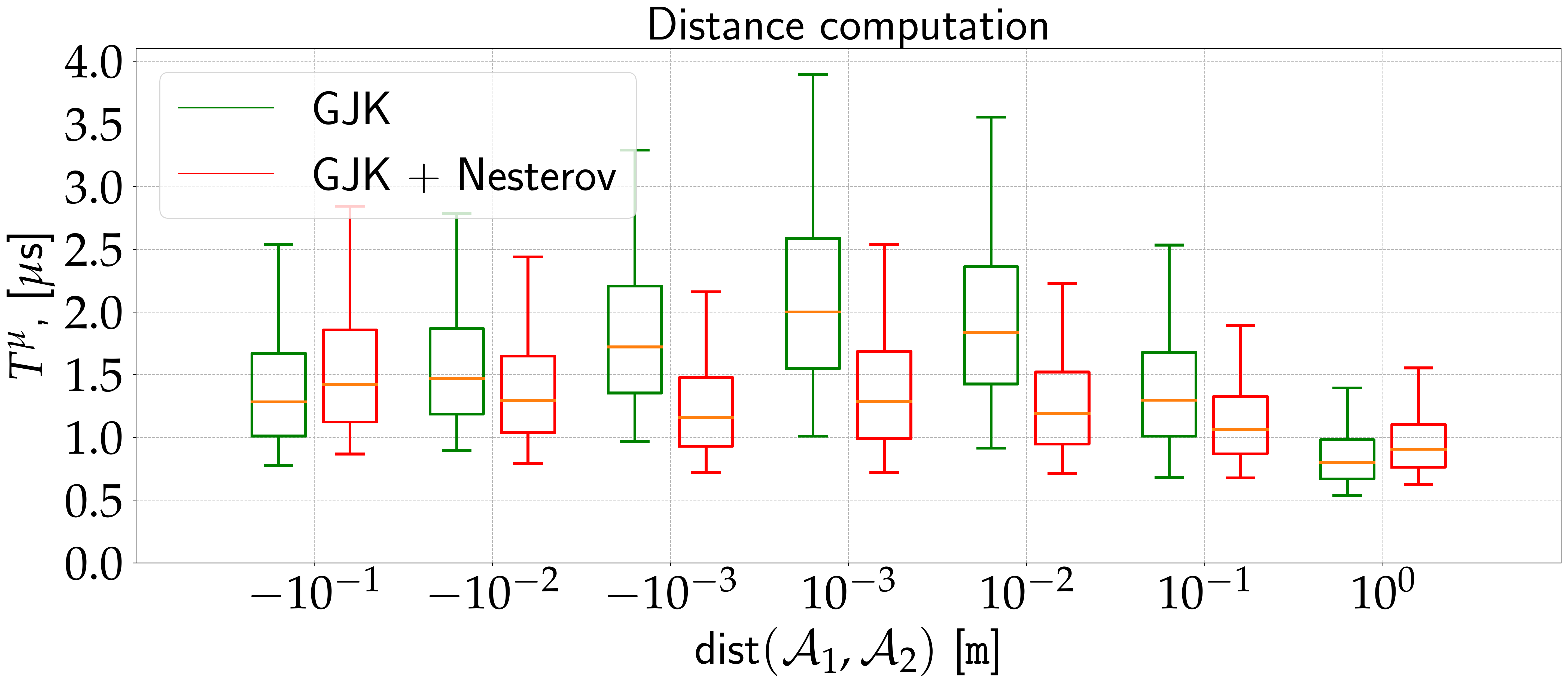}
    }
    \caption{\small
        \textbf{Distance computation on the ShapeNet benchmark.} The graphs show the number of iterations (a) and the execution time (b) for Nesterov-accelerated GJK with normalization (red) and vanilla GJK (green) for a range of distances (x-axis) between the shapes.
    }
    \label{fig:shapenet_benchmark_distances}
    \vspace{-0.5cm}
\end{figure}
%%% --> END FIGURE
%%% --> FIGURE
\begin{figure}
    % \vspace{2cm}
    \centering
    \subfloat[
        Number of iterations~$\numit$.
        Lower is better.
        \label{fig:shapenet_benchmark_distances_early_a}
    ]{
        \includegraphics[width=\linewidth]{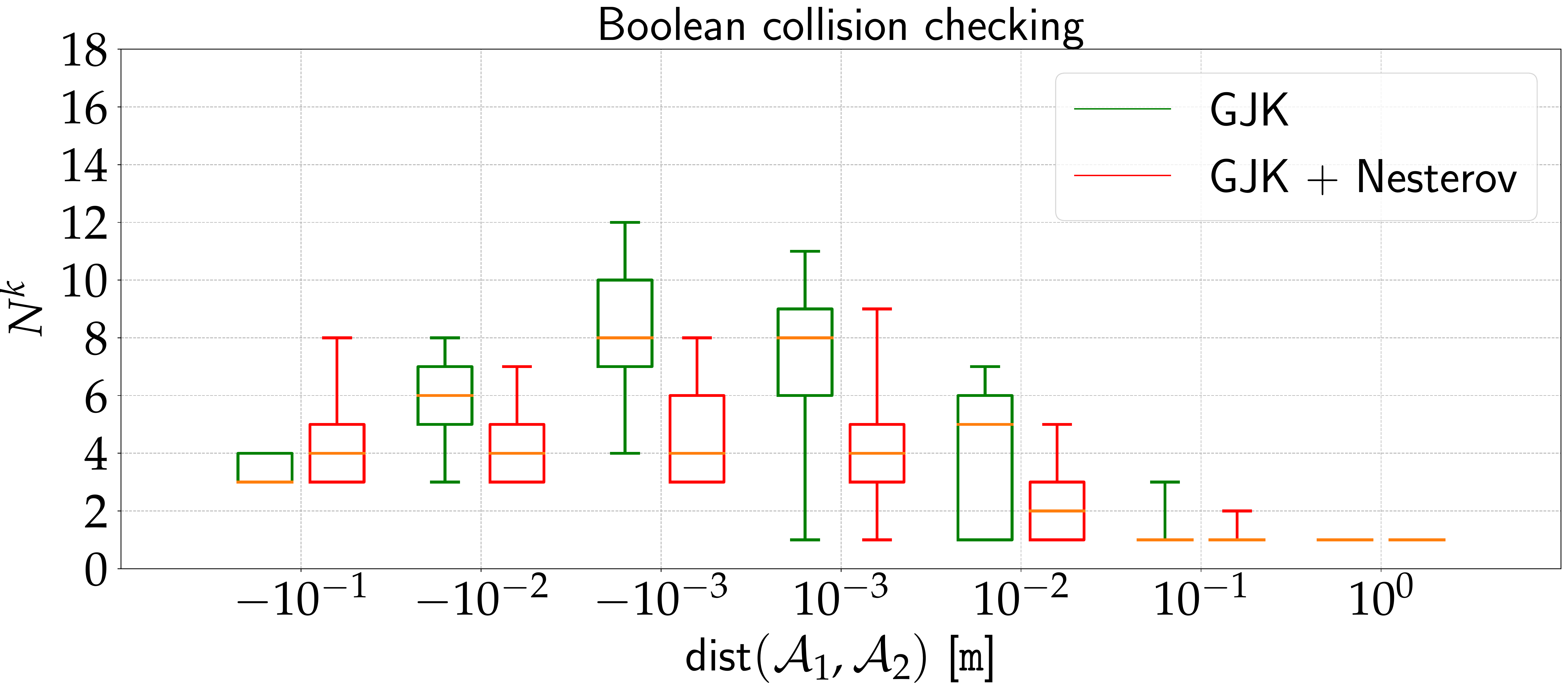}
    }\\
    \subfloat[
        Execution time~$\exectime$.
        Lower is better.
        \label{fig:shapenet_benchmark_distances_early_b}
    ]{
        \includegraphics[width=\linewidth]{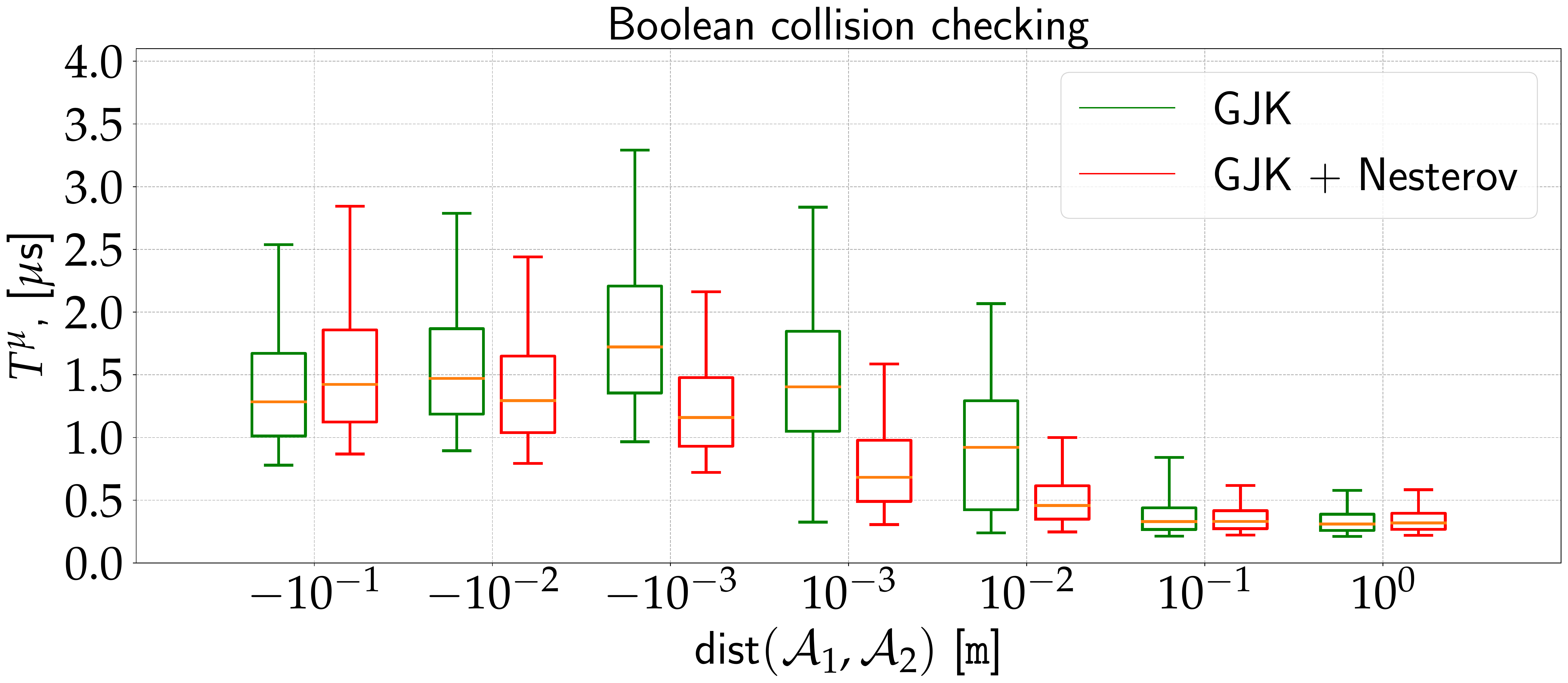}
    }
    \caption{\small
         \textbf{Boolean collision check on the ShapeNet benchmark.} The graphs show the number of iterations (a) and the execution time (b) for the Nesterov-accelerated GJK with normalization (red) and vanilla GJK (green) for a range of distances (x-axis) between the shapes.
    }
    \label{fig:shapenet_benchmark_distances_early}
    %\vspace{-0.5cm}
\end{figure}
%%% --> END FIGURE
%%% --> FIGURE
\begin{figure}
    % \vspace{2cm}
    \centering
    \subfloat[
        Ratio of number iteration~$\numitrel$. 
        Higher is better.
        \label{fig:shapenet_benchmark_distances_rel_a}
    ]{
        \includegraphics[width=\linewidth]{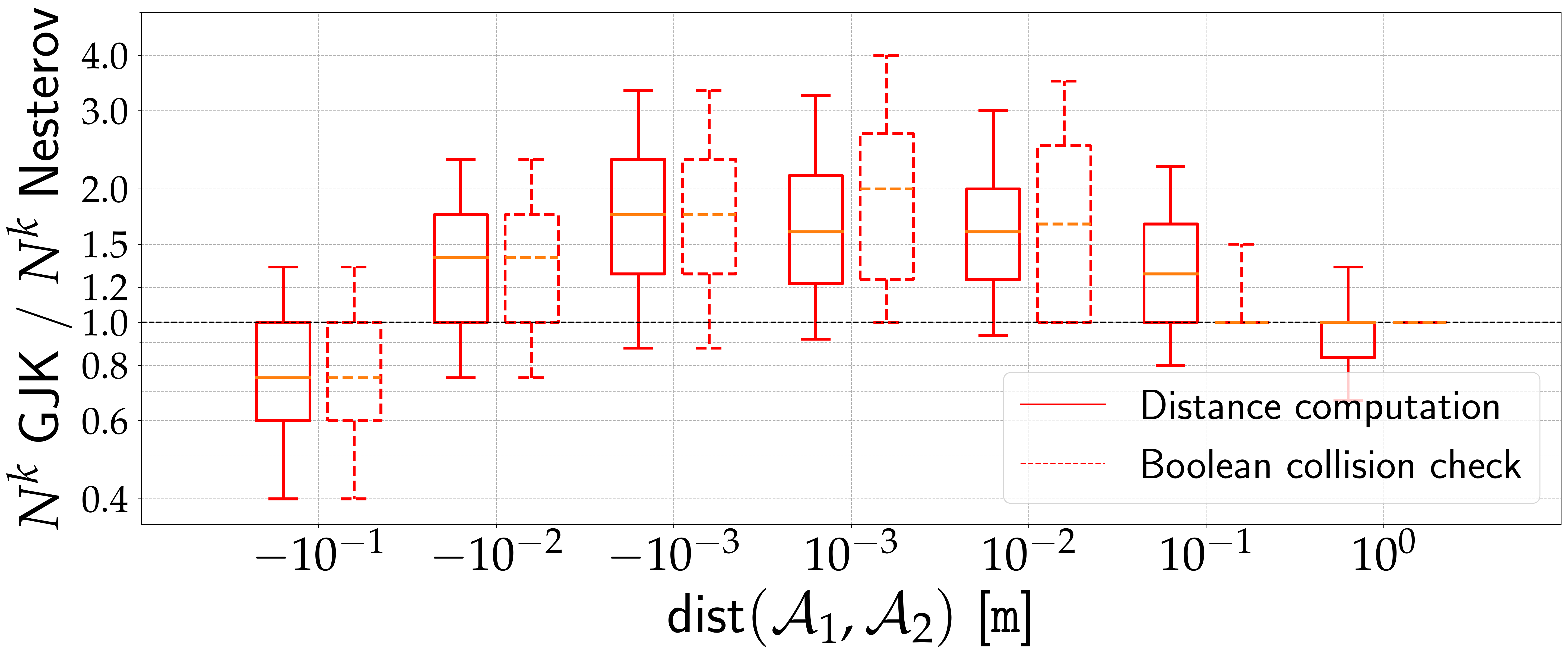}
    }\hfill\\
    \subfloat[
        Ratio of execution time~$\exectimerel$.
        Higher is better.
        \label{fig:shapenet_benchmark_distances_rel_b}
    ]{
        \includegraphics[width=\linewidth]{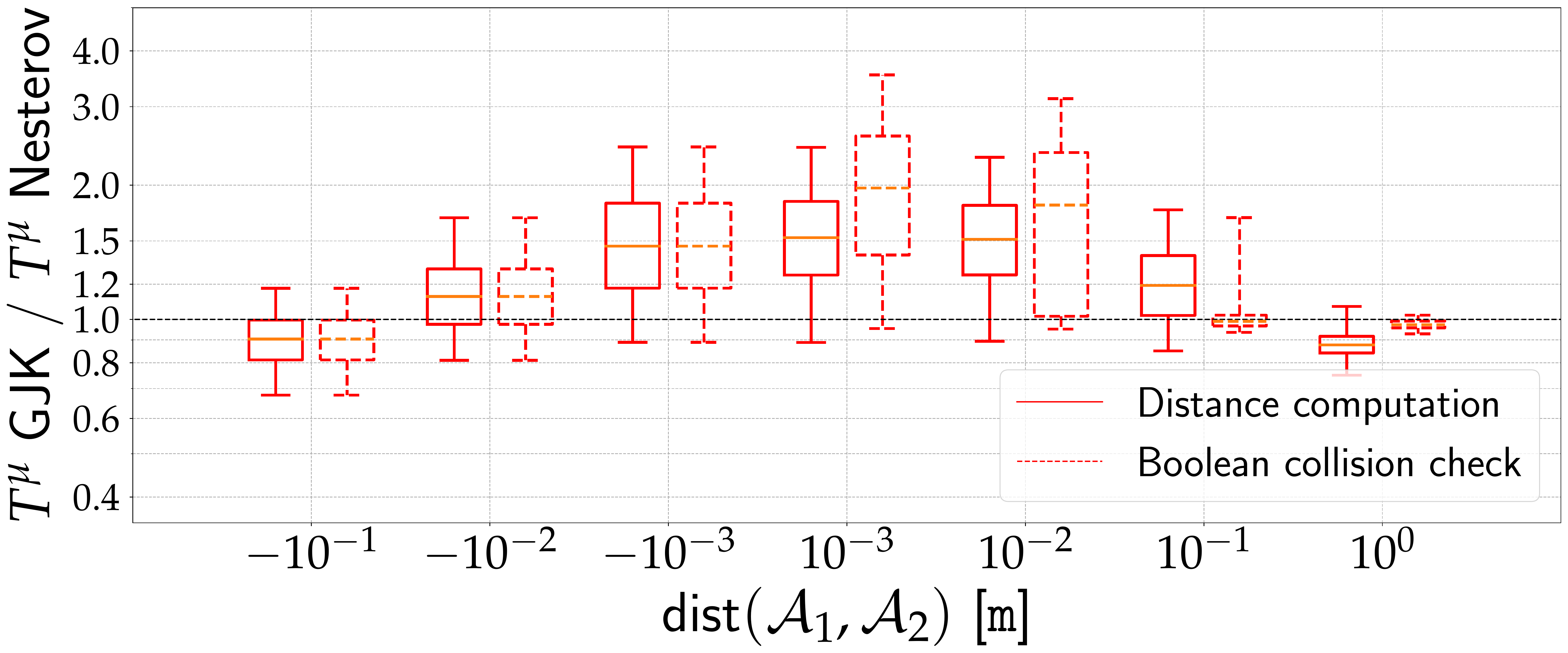}
    }
    \caption{\small 
        \textbf{Speed-ups on the ShapeNet benchmark.} The plots show ratios of the number of iterations (a) and execution times (b) of vanilla GJK and the Nesterov-accelerated GJK with normalization. Ratios over 1.0 show speed-ups of Nesterov-accelerated GJK over GJK. % for close-by and shallowly intersecting shapes.
    }
    \label{fig:shapenet_benchmark_distances_rel}
    %\vspace{-0.5cm}
\end{figure}
%%% --> END FIGURE

%%%%%%%%%% TABLE %%%%%%%%%
%%%%%%%%%%%%%%%%%%%%%%%%%%
\begin{table*}[!t]
    \renewcommand{\arraystretch}{2.0}
    \centering
    \begin{tabular}{cc|cc|cc|cc|cc|cc}
    %%%%%%%%%%
    % \hline
    \multicolumn{1}{c}{} &
    \multicolumn{1}{c}{} &
    \multicolumn{2}{c}{$N=33$} &
    \multicolumn{2}{c}{$N=141$} &
    \multicolumn{2}{c}{$N=626$} &
    \multicolumn{2}{c}{$N=2500$} \\
    \multicolumn{1}{c}{} &
    \multicolumn{1}{c}{} &
    \multicolumn{2}{c|}{\includegraphics[width=0.06\textwidth]{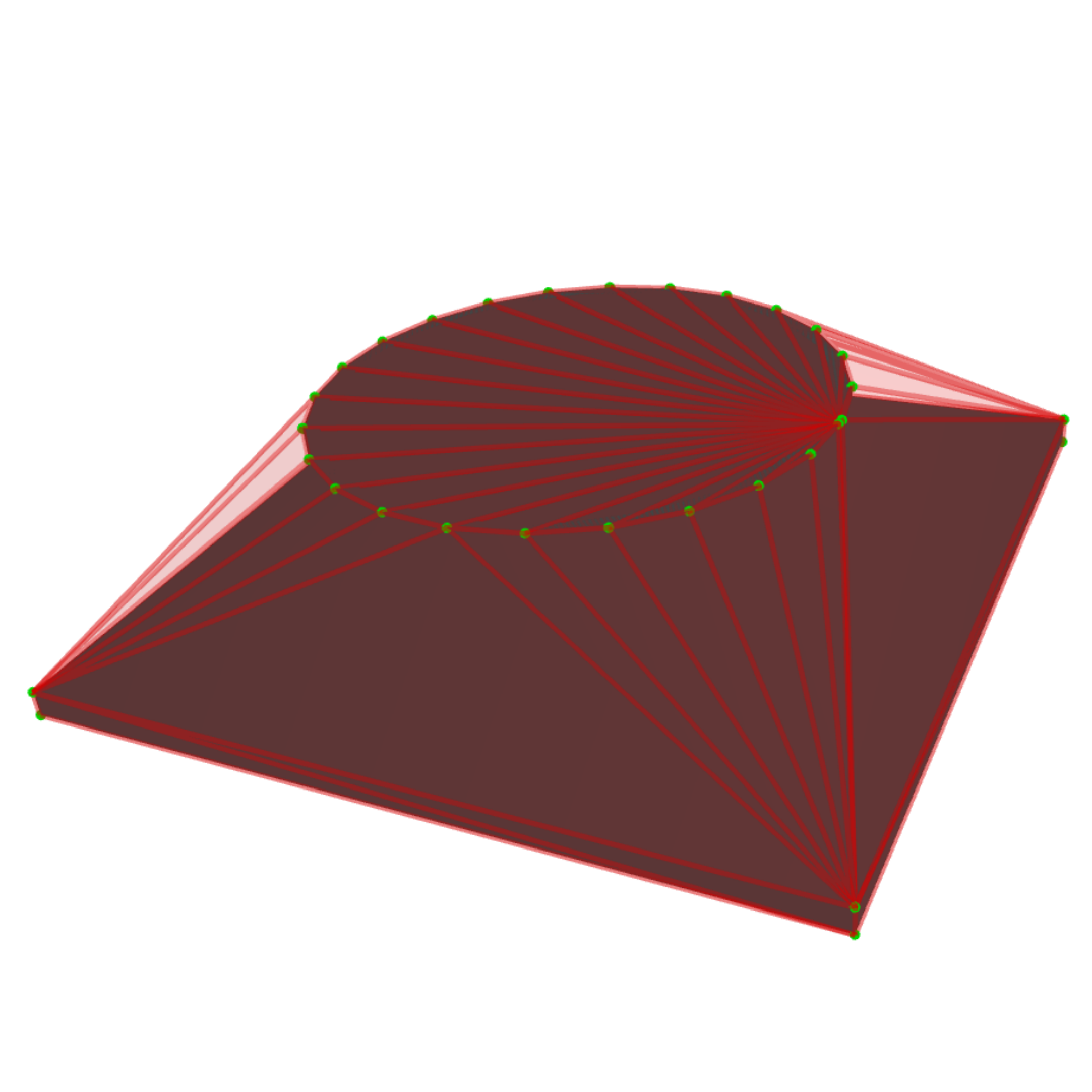}} &
    \multicolumn{2}{c|}{\includegraphics[width=0.06\textwidth]{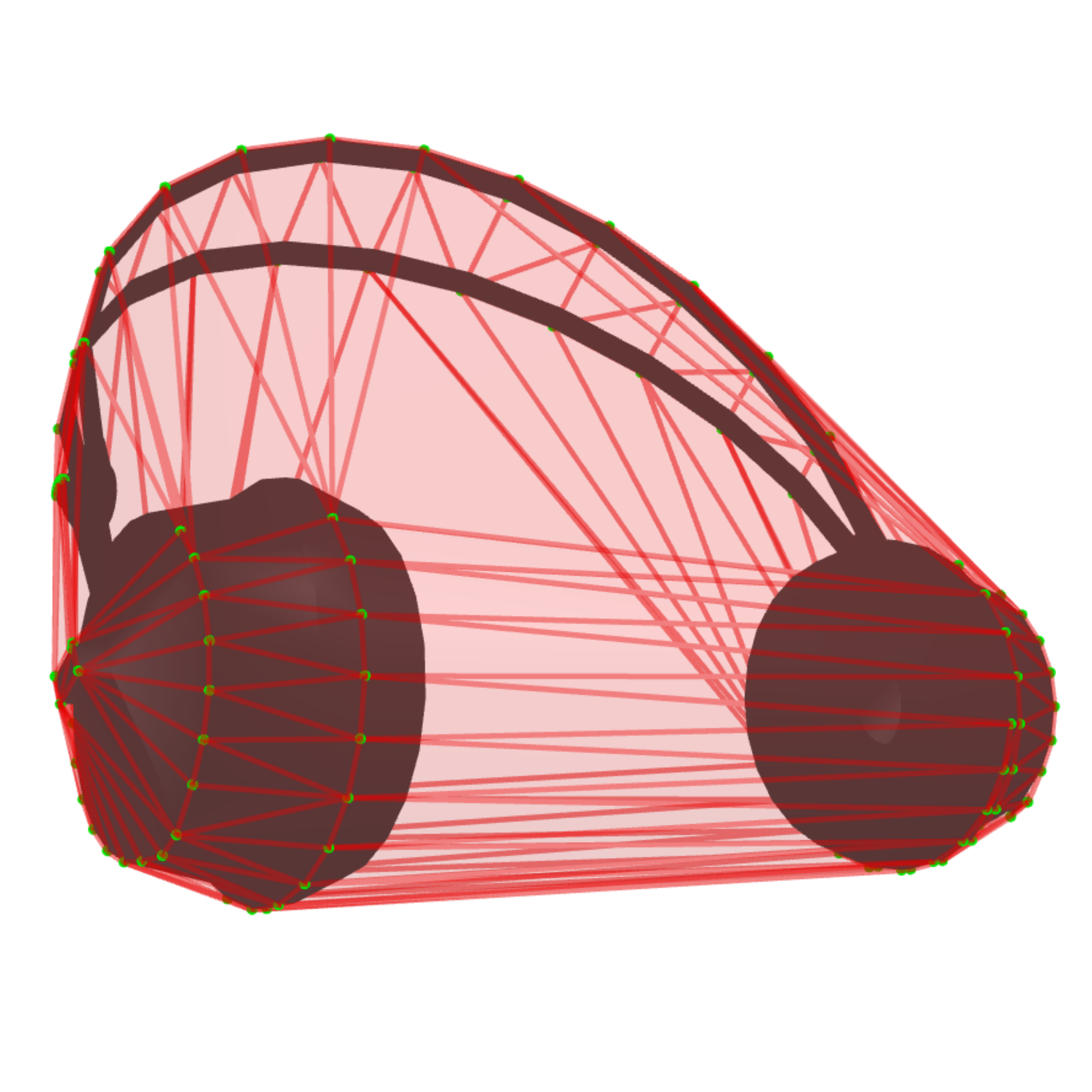}} &
    \multicolumn{2}{c|}{\includegraphics[width=0.06\textwidth]{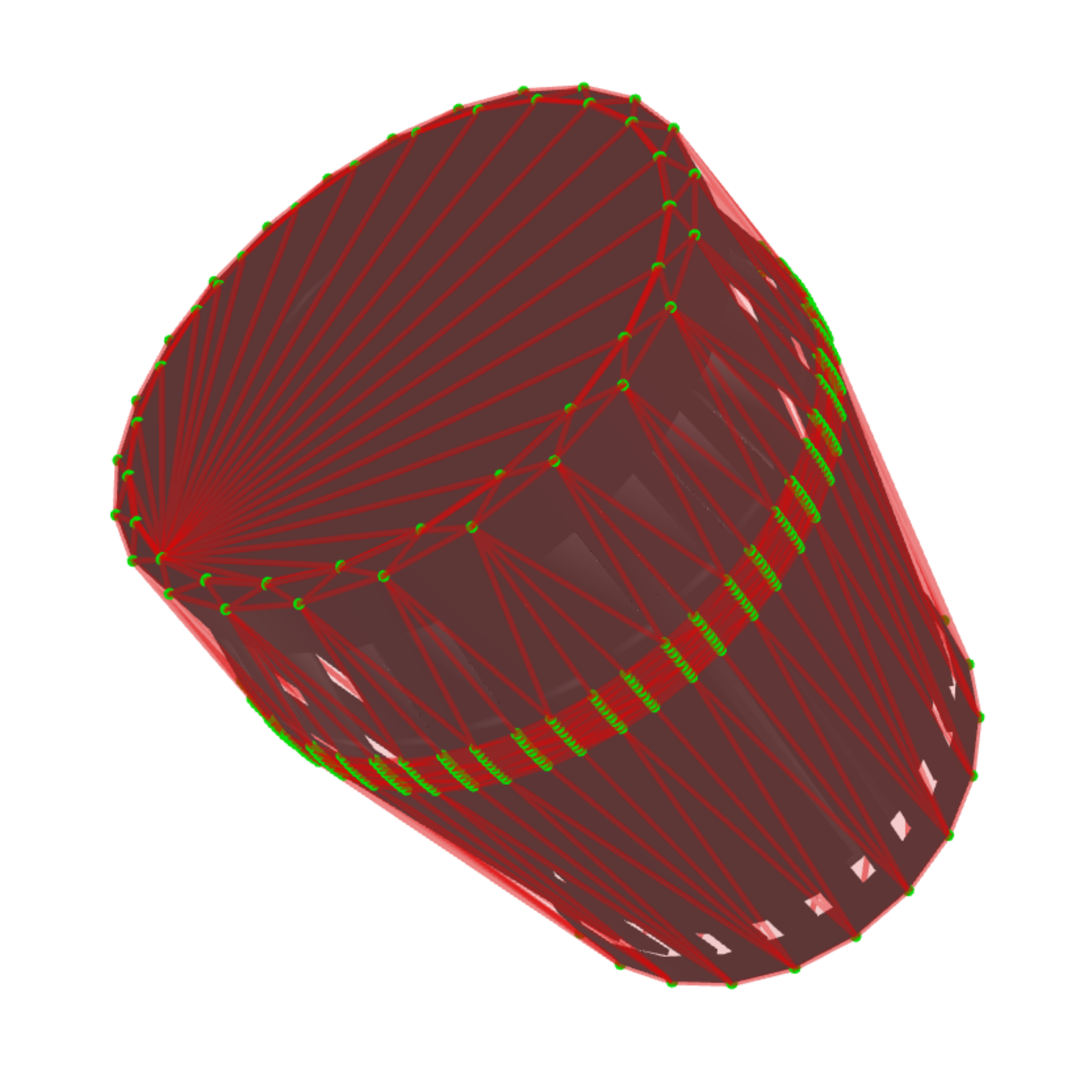}} &
    \multicolumn{2}{c}{\includegraphics[width=0.06\textwidth]{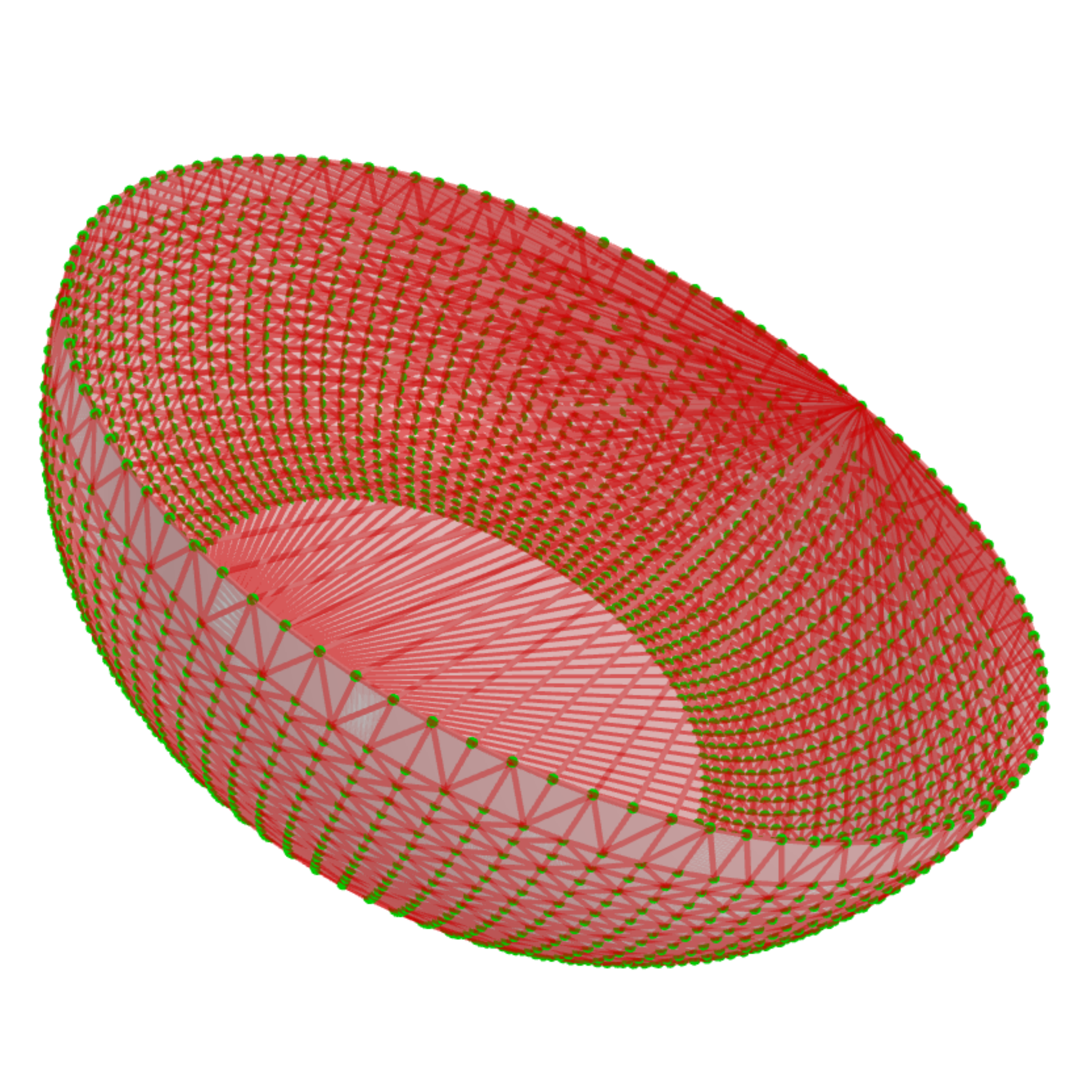}} \\
    %%%%%%%%%%
    % \hline
    %%%%%%%%%%
    \hline
    \multicolumn{1}{c}{} &
    \multicolumn{1}{c}{} &
    \multicolumn{1}{c}{\textbf{GJK}} &
    \multicolumn{1}{c|}{\textbf{GJK + Nesterov}} &
    \multicolumn{1}{c}{\textbf{GJK}} &
    \multicolumn{1}{c|}{\textbf{GJK + Nesterov}} &
    \multicolumn{1}{c}{\textbf{GJK}} &
    \multicolumn{1}{c|}{\textbf{GJK + Nesterov}} &
    \multicolumn{1}{c}{\textbf{GJK}} &
    \multicolumn{1}{c}{\textbf{GJK + Nesterov}} \\
    %%%%%%%%%%%%%%%%%%%%%%%%%%%%%%%
    %%%%%%%%%%%%%%%%%%%%%%%%%%%%%%%
    \hline
    %%%%%%%%%% OBJ 1 %%%%%%%%%%%%%
    \multirow{3}{*}{\vspace{5mm}\includegraphics[width=0.06\textwidth,height=0.06\linewidth]{figures/exp_2/obj1.pdf}} &
    \multicolumn{1}{c}{\exectimeD} &
    \multicolumn{1}{c}{$0.8 \pm 0.3$} &
    \multicolumn{1}{c|}{$0.8 \pm 0.2$} &
    
    \multicolumn{1}{c}{$1.1 \pm 0.4$} &
    \multicolumn{1}{c|}{$\bf{1.0 \pm 0.3}$} &
    
    \multicolumn{1}{c}{$1.3 \pm 0.4$} &
    \multicolumn{1}{c|}{$\bf{1.1 \pm 0.3}$} &
    
    \multicolumn{1}{c}{$2.5 \pm 0.6$} &
    \multicolumn{1}{c}{$\bf{1.8 \pm 0.6}$} \\
    
    \multicolumn{1}{c}{} &
    \multicolumn{1}{c}{\exectimeC} &
    \multicolumn{1}{c}{$0.6 \pm 0.3$} &
    \multicolumn{1}{c|}{$0.6 \pm 0.3$} &
    
    \multicolumn{1}{c}{$0.8 \pm 0.3$} &
    \multicolumn{1}{c|}{$\bf{0.7 \pm 0.3}$} &
    
    \multicolumn{1}{c}{$1.1 \pm 0.5$} &
    \multicolumn{1}{c|}{$\bf{0.8 \pm 0.4}$} &
    
    \multicolumn{1}{c}{$1.9 \pm 0.9$} &
    \multicolumn{1}{c}{$\bf{1.4 \pm 0.7}$} \\
    %%%%%%%%%% END OBJ 1 %%%%%%%%%%%%%
    %%%%%%%%%%%%%%%%%%%%%%%%%%%%%%%
    %%%%%%%%%%%%%%%%%%%%%%%%%%%%%%%
    \hline
    %%%%%%%%%% OBJ 2 %%%%%%%%%%%%%
    \multicolumn{1}{c}{\multirow{3}{*}{\vspace{5mm}\includegraphics[width=0.06\textwidth,height=0.06\linewidth]{figures/exp_2/obj2.pdf}}} &
    \multicolumn{1}{c}{\exectimeD} &
    \multicolumn{1}{c}{} &
    \multicolumn{1}{c|}{} &
    
    \multicolumn{1}{c}{$1.2\pm0.3$} &
    \multicolumn{1}{c|}{$\bf{1.0\pm0.2}$} &
    
    \multicolumn{1}{c}{$1.5\pm0.4$} &
    \multicolumn{1}{c|}{$\bf{1.2\pm0.3}$} &
    
    \multicolumn{1}{c}{$2.6\pm0.7$} &
    \multicolumn{1}{c}{$\bf{1.9\pm0.6}$} \\
    
    \multicolumn{1}{c}{} & 
    \multicolumn{1}{c}{\exectimeC} &
    \multicolumn{1}{c}{} & 
    \multicolumn{1}{c|}{} &
    
    \multicolumn{1}{c}{$0.9\pm0.4$} &
    \multicolumn{1}{c|}{$\bf{0.7\pm0.3}$} &
    
    \multicolumn{1}{c}{$1.2\pm0.5$} &
    \multicolumn{1}{c|}{$\bf{0.9\pm0.4}$} &
    
    \multicolumn{1}{c}{$2.1\pm0.9$} &
    \multicolumn{1}{c}{$\bf{1.4\pm0.7}$} \\
    %%%%%%%%%% END OBJ 2 %%%%%%%%%%%%%
    %%%%%%%%%%%%%%%%%%%%%%%%%%%%%%%
    %%%%%%%%%%%%%%%%%%%%%%%%%%%%%%
    \hline
    %%%%%%%%%% OBJ 3 %%%%%%%%%%%%%
    \multicolumn{1}{c}{\multirow{3}{*}{\vspace{5mm}\includegraphics[width=0.06\textwidth,height=0.06\linewidth]{figures/exp_2/obj3.pdf}}} &
    \multicolumn{1}{c}{\exectimeD} &
    \multicolumn{1}{c}{} &
    \multicolumn{1}{c|}{} &
    
    \multicolumn{1}{c}{} &
    \multicolumn{1}{c|}{} &
    
    \multicolumn{1}{c}{$1.8\pm0.6$} &
    \multicolumn{1}{c|}{$\bf{1.4\pm0.4}$} &
    
    \multicolumn{1}{c}{$3.0\pm0.8$} &
    \multicolumn{1}{c}{$\bf{2.1\pm0.7}$} \\
    
    \multicolumn{1}{c}{} &
    \multicolumn{1}{c}{\exectimeC} &
    \multicolumn{1}{c}{} &
    \multicolumn{1}{c|}{} &
    
    \multicolumn{1}{c}{} &
    \multicolumn{1}{c|}{} &
    
    \multicolumn{1}{c}{$1.4\pm0.6$} &
    \multicolumn{1}{c|}{$\bf{1.0\pm0.5}$} &
    
    \multicolumn{1}{c}{$2.5\pm1.1$} &
    \multicolumn{1}{c}{$\bf{1.6\pm0.8}$} \\
    %%%%%%%%%% END OBJ 3 %%%%%%%%%%%%%
    %%%%%%%%%%%%%%%%%%%%%%%%%%%%%%%
    %%%%%%%%%%%%%%%%%%%%%%%%%%%%%%%
    \hline
    %%%%%%%%%% OBJ 4 %%%%%%%%%%%%%
    \multicolumn{1}{c}{\multirow{3}{*}{\vspace{5mm}\includegraphics[width=0.06\textwidth,height=0.06\linewidth]{figures/exp_2/obj4.pdf}}} &
    \multicolumn{1}{c}{\exectimeD} &
    \multicolumn{1}{c}{} &
    \multicolumn{1}{c|}{} &
    
    \multicolumn{1}{c}{} &
    \multicolumn{1}{c|}{} &
    
    \multicolumn{1}{c}{} &
    \multicolumn{1}{c|}{} &
    
    \multicolumn{1}{c}{$4.1\pm1.1$} &
    \multicolumn{1}{c}{$\bf{2.7\pm0.9}$} \\
    
    \multicolumn{1}{c}{} &
    \multicolumn{1}{c}{\exectimeC} &
    \multicolumn{1}{c}{} &
    \multicolumn{1}{c|}{} &
    
    \multicolumn{1}{c}{} &
    \multicolumn{1}{c|}{} &
    
    \multicolumn{1}{c}{} &
    \multicolumn{1}{c|}{} &
    
    \multicolumn{1}{c}{$3.3\pm1.5$} &
    \multicolumn{1}{c}{$\bf{\bf{2.2\pm1.1}}$} \\
    %%%%%%%%%% END OBJ 4 %%%%%%%%%%%%%
    \end{tabular}
    \caption{\small
        \textbf{Computation times ($\mu s$) for distance computation (\exectimeD) and Boolean collision checking (\exectimeC) on the ShapeNet benchmark} 
        for close-by or shallowly intersecting shapes where the distance between shapes is~\mbox{$-0.1\meter \leq \dshapes \leq 0.1\meter$}. $N$ denotes the number of vertices for each mesh.
        For all depicted pairs, Nesterov-accelerated GJK always gives a better (or at least as good) timing for both collision computation and distance evaluation.
    }
    \label{tab:shapenet_table}
\end{table*}
\vspace{0.2cm}
\noindent
\textbf{Statistical validation over the ShapeNet dataset.}
In Fig.~\ref{fig:shapenet_benchmark_distances} and Fig.~\ref{fig:shapenet_benchmark_distances_early}, we report the number of iterations~\mbox{$\numit$} and execution time~\mbox{$\exectime$} for both Nesterov-accelerated and vanilla GJK.
In Fig.~\ref{fig:shapenet_benchmark_distances_rel}, we report relative accelerations~\mbox{$\numitrel$} and~\mbox{$\exectimerel$} of Nesterov-accelerated GJK compared to GJK on a given collision problem.
These relative measures allow to analyze the effects of the two algorithms on exactly the same collision problems, which are not captured when using absolute values.  
Overall, Nesterov-accelerated GJK significantly reduces the number of iterations and execution time when compared to GJK in cases where shapes are shallowly intersecting or in close-proximity.
The number of iterations and the execution times are reduced by a factor ranging from~$1.5$ up to~$2$.
It is worth recalling at this stage, that when two shapes are relatively far from each other, any broad phase algorithm will automatically discard such pair. 
Only in a small percentage of cases Nesterov-accelerated GJK is slower than GJK.
Momentum restarts could help improve these rare cases and will be considered in future work.

In Table.~\ref{tab:shapenet_table}, we select 4 meshes with an increasing number of vertices to highlight the benefits of the Nesterov acceleration.
For each pair, we report the mean and the standard deviation of the execution time for both distance computation and Boolean collision checking. 
We consider the challenging set-up of close-by or shallowly intersecting shapes in the range of separation distances \mbox{$-0.1\,\meter \leq \dshapes \leq  0.1\,\meter$}.
The lower mean and standard deviation show that Nesterov-accelerated GJK is faster than the vanilla GJK and reduces the spread of computation times across the different collision problems in this setting.

\subsection{Importance of the simplex strategy in GJK}
\noindent
In the previous sections, we have experimentally shown the improvement of our method, Nesterov-accelerated GJK, over the vanilla GJK algorithm for collision problems which are important in practice, \textit{i.e.} when collision pairs have not been filtered by the broad phase and are thus overlapping or in close-proximity.
We conclude this section by demonstrating the importance of the simplex strategy used in GJK and our method when solving collision problems.
To do so, we evaluate the performance of the Frank-Wolfe algorithm (Alg.~\ref{alg:FW_vanilla_linesearch}), the recent NESMINO algorithm~\cite{nesmino2019}, GJK and our method on ellipsoids and cubes and report the results in Table.~\ref{tab:nesmino_comparison}.
Altough the NESMINO algorithm is similar to projected-gradient descent and strongly differs from Frank-Wolfe like algorithms, it uses the classic Nesterov acceleration which makes it interesting to compare to our method.
FW, GJK and our method stop when a tolerance of~$\epsilon = 10^{-8}$ on the FW duality-gap is met.
Therefore, to render the NESMINO algorithm comparable to the other considered methods, we run NESMINO until the distance between its solution and the solution found by GJK is less than~$\sqrt{\epsilon} = 10^{-4}$.

In Table~\ref{tab:nesmino_comparison_ellipsoids}, we consider 1000 collision problems between pairs of ellipsoids for each distance category (overlapping, close-proximity and distant).
Shapes are in close-proximity when~\mbox{$0 \leq \dshapes \leq 0.1 \, \meter$}.
In this first scenario, all algorithms have a comparable number of operations per iteration.
Indeed, the projection operation used in NESMINO when the shapes are ellipsoids has the same complexity as the support operation used in the three other algorithms.
Although GJK and our method also do a simplex projection at each iteration, this operation has about the same complexity as computing the support point.
In the case of strictly-convex shapes such as ellipsoids, GJK and our method significantly outperform the FW and NESMINO algorithms.
This is especially the case when the shapes are overlapping or in close-proximity where GJK or our method take 3 to 10 times less iterations compared to FW or NESMINO.

In Table~\ref{tab:nesmino_comparison_cubes}, we repeated the same experiments with collision pairs of cubes.
Since cubes are polytopes with a small number of vertices, GJK and our method only take a few iterations to reach a tolerance of~$\epsilon=10^{-8}$.
However, because cubes are non-strictly convex shapes, the convergence of the FW algorithm is~$O(1/\epsilon)$, \textit{i.e.} it takes on the order of~$1/\epsilon$ iterations to reach a FW duality-gap of~$\epsilon$.
The NESMINO algorithm takes less iterations than FW but more than 100 times more iterations than GJK and our method.
In the specific case of polytopes, the NESMINO algorithm is also much more costly per iteration than FW, GJK or our method, as it replaces the computation of support points by much more costly projections on the original polytopes.
\begin{table}[!t]
    \centering
    \subfloat[Distance computation - Ellipsoids.
    \label{tab:nesmino_comparison_ellipsoids}]{
    \begin{tabular}{ c c c c c }
      & FW & Nesmino & GJK & Ours \\ 
      & (Alg.~\ref{alg:FW_vanilla_linesearch}) & \cite{nesmino2019} & (Alg.~\ref{alg:FW_fully_corrective_simplex_GJK}) & (Alg.~\ref{alg:GJK_nesterov_algo}) \\ 
      \hline
     Overlapping
        &  $73 \pm 62$ & $50 \pm 18$ & $\bf{6 \pm 2}$ & $\bf{6 \pm 3}$ \\ 
      \hline
     Close-proximity
        &  $48 \pm 42$ & $74 \pm 29$ & $16 \pm 5$ & $\bf{7 \pm 2}$ \\ 
      \hline
     Distant
        & $\bf{4 \pm 1}$ & $18 \pm 2$ & $\bf{4 \pm 1}$ & $13 \pm 4$ \\ 
    \end{tabular}
    }\\
    \subfloat[Distance computation - Cubes.
    \label{tab:nesmino_comparison_cubes}]{
    \begin{tabular}{ c c c c c }
      & FW & Nesmino & GJK & Ours \\ 
      & (Alg.~\ref{alg:FW_vanilla_linesearch}) & \cite{nesmino2019} & (Alg.~\ref{alg:FW_fully_corrective_simplex_GJK}) & (Alg.~\ref{alg:GJK_nesterov_algo}) \\ 
      \hline
     Overlapping
        & $5.4\text{k} \pm 4.8\text{k}$ & $922 \pm 244$ & $6 \pm 1$ & $\bf{5 \pm 1}$ \\ 
      \hline
     Close-proximity
        & $14.3\text{k} \pm 9.7\text{k}$ & $828 \pm 225$ & $5 \pm 1$ & $\bf{4 \pm 1}$ \\ 
      \hline
     Distant
        & $13.1\text{k} \pm 13.5\text{k}$ & $623 \pm 219$ & $4 \pm 1$ & $\bf{3 \pm 1}$ \\ 
    \end{tabular}
    }
    \caption{\small \textbf{Number of iterations for distance computation between ellipsoids (a) and between cubes (b).}
    For each distance category (overlapping, close-proximity and distant), we report the mean and standard deviation of the number of iterations. 
    Shapes are in close-proximity when~\mbox{$0 \leq \dshapes \leq 0.1 \, \meter$}.
    }
    \label{tab:nesmino_comparison}
\end{table}

% --> END SECTION IV

% CONCLUSION
\section{Conclusion} 
\label{sec:conclusion}
In this work, we have first established that the well-known GJK algorithm can be understood as a variant of the Frank-Wolfe method, well studied within the convex optimization community, and more precisely GJK can be identified as a sub-case of fully-corrective Frank-Wolfe. 
Subsequently, this connection has enabled us to accelerate the GJK algorithm in the sense of Nesterov acceleration, by adapting recent contributions on the application of Nesterov acceleration to the context of Frank-Wolfe.
Through extensive benchmarks, we have shown that this acceleration is beneficial for both collision detection and distance computation settings for scenarios where shapes are intersecting or in close proximity, accelerating collision detection by up to a factor of 2.
Interestingly, these two scenarios notably encompass the generic contexts of planning and control as well as physical simulation, which are essential areas of modern robotics.
Therefore, although the proposed acceleration corresponds to improvements of GJK's execution time on the order of few microseconds, modern robotics applications may solve millions to billions of collision problems \textit{e.g.} when learning a policy with RL~\cite{openaiSolvingRubikCube2019}.
The Nesterov acceleration of GJK is already included in the HPP-FCL library~\cite{hppfclweb}, notably used by the HPP framework~\cite{mirabel2016hpp} for motion planning and the Pinocchio framework~\cite{carpentier2019pinocchio} dedicated to simulation and modelling.
One can expect this work to be largely adopted in the current available GJK implementations, as it only requires minor algorithmic changes.
This work should benefit a large audience within robotics (e.g., simulation, planning, control) and beyond by addressing issues shared by other communities, including computer graphics and computational geometry.

\section*{Acknowledgments}
The authors would like to warmly thank Francis Bach for fruitful discussions on Min-Norm Point algorithms.
This work was partly supported by the European Regional Development Fund under the project IMPACT (reg. no. CZ.02.1.01/0.0/0.0/15 003/0000468), by the French government under management of Agence Nationale de la Recherche as part of the “Investissements d’avenir” program, reference ANR-19-P3IA-0001 (PRAIRIE 3IA Institute) and the Louis Vuitton ENS Chair on Artificial Intelligence.

%% Use plainnat to work nicely with natbib. 

%\newpage
\balance{}
\bibliographystyle{plainnat}
\bibliography{references}

\end{document}